\documentclass[12pt,preprint]{elsarticle}

\usepackage[utf8]{inputenc}

\usepackage{fullpage}

\usepackage[british]{babel}
\usepackage{caption}
\usepackage{commath}
\usepackage{amsmath}
\usepackage{amssymb}
\usepackage{hyperref}
\usepackage[capitalise]{cleveref}
\usepackage{graphicx}
\graphicspath{{figures/}}
\usepackage{url}

\usepackage[dvipsnames]{xcolor}
\hypersetup{
    colorlinks=true,
    linkcolor={red!50!black!50!},
    citecolor={blue!50!black!50!},
    urlcolor={blue!80!black!20!}
}

\DeclareMathOperator*{\argmax}{arg\,max}
\DeclareMathOperator*{\argmin}{arg\,min}

\crefformat{equation}{Eq.~#2#1#3}
\DeclareRobustCommand{\pcref}[1]{%
  \begingroup
  \cref{#1}%
  \endgroup
}

\usepackage[mathlines]{lineno}

\usepackage{color}
\definecolor{Blue}{RGB}{0,0,255}
\definecolor{Red}{RGB}{255,0,0}

\newcommand{\chg}[2]{{#1} {#2}} 

\begin{document}

% --------------------------------------------------------------------------------
% Frontmatter
% --------------------------------------------------------------------------------

\begin{frontmatter}

\title{An empirical evaluation of active inference in multi-armed bandits}

\author[affil1,affil2]{Dimitrije Markovi\'c\corref{mycorrespondingauthor}\fnref{myfootnote}}
\ead{dimitrije.markovic@tu-dresden.de}
\author[affil4,affil5]{Hrvoje Stoji\'c\fnref{myfootnote}}
\author[affil1]{Sarah Schw\"obel}
\author[affil1,affil2]{Stefan J. Kiebel}

\address[affil1]{Faculty of Psychology, Technische Universit\"at Dresden, 01062 Dresden, Germany}
\address[affil2]{Centre for Tactile Internet with Human-in-the-Loop (CeTI), \\Technische Universit\"at Dresden, 01062 Dresden, Germany}
\address[affil4]{Max Planck UCL Centre for Computational Psychiatry and Ageing Research, University College London,
10-12 Russell Square, London, WC1B 5EH, United Kingdom}
\address[affil5]{Secondmind, 72 Hills Rd, Cambridge, CB2 1LA, United Kingdom}
\fntext[myfootnote]{These authors contributed equally}
\cortext[mycorrespondingauthor]{Corresponding author}

%Footnotes should be used sparingly. Number them consecutively throughout the article. Many word processors can build footnotes into the text, and this feature may be used. Otherwise, please indicate the position of footnotes in the text and list the footnotes themselves separately at the end of the article. Do not include footnotes in the Reference list.

\begin{abstract} % A concise and factual abstract is require which should not exceed 250 words .
A key feature of sequential decision making under uncertainty is a need to balance between exploiting--choosing the best action according to the current knowledge, and exploring--obtaining information about values of other actions. The multi-armed bandit problem, a classical task that captures this trade-off, served as a vehicle in machine learning for developing bandit algorithms that proved to be useful in numerous industrial applications. The active inference framework, an approach to sequential decision making recently developed in neuroscience for understanding human and animal behaviour, is distinguished by its sophisticated strategy for resolving the exploration-exploitation trade-off. This makes active inference an exciting alternative to already established bandit algorithms. Here we derive an efficient and scalable approximate active inference algorithm and compare it to two state-of-the-art bandit algorithms: Bayesian upper confidence bound and optimistic Thompson sampling. This comparison is done on two types of bandit problems: a stationary and a dynamic switching bandit. Our empirical evaluation shows that the active inference algorithm does not produce efficient long-term behaviour in stationary bandits. However, in the more challenging switching bandit problem active inference performs substantially better than the two state-of-the-art bandit algorithms. The results open exciting venues for further research in theoretical and applied machine learning, as well as lend additional credibility to active inference as a general framework for studying human and animal behaviour.
% 229 words
\end{abstract}


\end{frontmatter}

% --------------------------------------------------------------------------------
% Intro
% --------------------------------------------------------------------------------

\section{Introduction}
\setcounter{figure}{0}

When we are repeatedly deciding between alternative courses of action -- about whose outcomes we are uncertain -- we have to strike a trade-off between exploration and exploitation. Do we exploit and choose an option that we currently expect to be the best, or do we sample more options with uncertain outcomes in order to learn about them, and potentially find a better option? This trade-off is one of the fundamental problems of sequential decision making and it has been extensively studied both in the context of neuroscience \cite{wilson2020balancing,mehlhorn2015unpacking,cohen2007should} as well as machine learning \cite{lattimore2020bandit, chapelle2011empirical, auer2002finite,kaufmann2018bayesian}. Here we propose active inference -- an approach to sequential decision making developed recently in neuroscience  \cite{kaplan2017navig, friston2017temporal, friston2017process, mirza2016scene, friston2016learning} -- as an attractive alternative to established algorithms in machine learning. Although the exploration-exploitation trade-off has been described and analysed within the active inference framework \cite{fitzgerald2015active,friston2015active,schwartenbeck2013exploration}, the focus was on explaining animal and human behaviour rather than the algorithm performance on a given problem. What is lacking for a convincing machine learning application is the evaluation on multi-armed bandit problems \cite{lattimore2020bandit}, a set of standard problems that isolate the exploration-exploitation trade-off, thereby enabling a focus on best possible performance and the comparison to state-of-the-art bandit algorithms from machine learning. Conversely, these analyses will also feed back into neuroscience research, giving rational foundations to active inference explanations of animal and human behaviour.

When investigating human and animal behaviour in stochastic (uncertain) environments, it has become increasingly fruitful to model and describe behaviour based on principles of Bayesian inference \cite{friston2012history,doya2007bayesian,knill2004bayesian}, both when describing perception, and decision making and planning \cite{botvinick2012planning}. The approach to describing sequential decision making and planning as probabilistic inference is jointly integrated within active inference \cite{kaplan2017navig, friston2017temporal, friston2017process, mirza2016scene,friston2016learning,schwartenbeck2013exploration}, a mathematical framework for solving partially observable Markov decision processes, derived from the general self-organising principle for biological systems -- the free energy principle \cite{karl2012free,friston2006free}. Recent work has demonstrated that different types of exploratory behaviour -- directed and random exploration -- naturally emerge within active inference \cite{schwartenbeck2019computational}. This makes active inference a useful approach for modelling how animals and humans resolve the exploration-exploitation trade-off, but also points at its potential usefulness for bandit and reinforcement learning problems in machine learning where the exploration-exploitation trade-off plays a prominent role \cite{lattimore2020bandit,sutton2018reinforcement}. Active inference in its initial form was developed for small state spaces and toy problems without consideration for applications to typical machine learning problems. This has recently changed and various scalable solutions have been proposed \cite{ueltzhoffer2018deep,millidge2020deep}, in addition to complex sequential policy optimisation that involves sophisticated (deep tree) searches \cite{friston2020sophisticated,fountas2020deep}. Therefore, to make the active inference approach practical and scalable to bandit problems typically used in machine learning, we introduce here an approximate active inference (A-AI) algorithm.

Here we examine how well the exact and A-AI algorithms perform in multi-armed-bandit problems that are traditionally used as benchmarks in the research on the exploration-exploitation trade-off \cite{lattimore2020bandit}. Although originally formulated for improving medical trials \cite{thompson1933likelihood}, multi-armed bandits have become an essential tool for studying human learning and decision making early on \cite{bush1953stochastic}, and later on attracted the attention of statisticians \cite{whittle1980multi,lai1985asymptotically} and machine learning researchers \cite{lattimore2020bandit} for studying the nature of sequential decision making more generally. We consider two types of bandit problems in our empirical evaluation: a stationary bandit as a classical machine learning problem \cite{lai1985asymptotically,auer2002finite,kaufmann2018bayesian,kaufmann2012thompson} and a switching bandit commonly used in neuroscience \cite{izquierdo2017neural,markovic2019predicting,iglesias2013hierarchical,wilson2012inferring,racey2011pigeon,behrens2007learning}. This will make the presented results directly relevant not only for the machine learning community, but also for learning and decision making studies in neuroscience, which are often utilising the active inference framework for a wide range of research questions. 

Using these two types of bandit problems we empirically compare the active inference algorithm to two state-of-the-art bandit algorithms from machine learning: a variant of the upper confidence bound (UCB) \cite{auer2002finite} algorithm -- the Bayesian UCB algorithm \cite{kaufmann2012bayesian,kaufmann2012thompson,kaufmann2018bayesian} -- and a variant of Thompson sampling -- optimistic Thompson sampling \cite{lu2019adaptive}. Both types of algorithms keep track of uncertainty about the values of actions, in the form of posterior beliefs about reward probabilities, and leverage these to balance between exploration and exploitation, albeit in a different way. These two algorithms reach state-of-the-art performance on various types of stationary bandit problems \cite{auer2002finite,chapelle2011empirical,kaufmann2012bayesian,kaufmann2012thompson}, achieving regret (the difference between actual and optimal performance) that is close to the best possible logarithmic regret \cite{lai1985asymptotically}. In switching bandits, learning is more complex, but once this is properly accounted for, both the optimistic Thompson sampling and Bayesian UCB exhibit the state-of-the-art performance \cite{cao2018nearly,alami2017memory,russo2018tutorial,lu2019adaptive,roijers2017interactive}.

We use a Bayesian approach to the bandit problem, also known as Bayesian bandits \cite{wang1992bayesian}, for all algorithms -- active inference, Bayesian UCB and optimistic Thompson sampling. The Bayesian treatment allows us to keep the learning rules equivalent, thus facilitating the comparison of different action selection strategies. In other words, belief updating and learning of the hidden reward probabilities exclusively rests on the learning rules derived from an (approximate) inference scheme, and are independent on the specific action selection principle \cite{lu2019adaptive}. Furthermore, learning algorithms derived from principles of Bayesian inference can be made domain-agnostic and fully adaptive to a wide range of unknown properties of the underlying bandit dynamics, such as the frequency of changes of choice-reward contingencies. Therefore, we use the same inference scheme for all algorithms -- variational surprise minimisation learning (SMiLE), an algorithm inspired by recent work in the field of human and animal decision making in changing environments \cite{liakoni2021learning,markovic2016comparative}. The variational SMiLE algorithm corresponds to online Bayesian inference modulated by surprise, which can be expressed in terms of simple delta-like learning rules operating on the sufficient statistics of posterior beliefs. 

In what follows, we will first introduce in detail the two types of bandit problems we focus on: the stationary and the dynamic bandit problem. We first describe each bandit problem formally in an abstract way and then specify the particular instantiation we use in our computational experiments. We will constrain ourselves to a well-studied version of bandits, the so-called Bernoulli bandits. For Bernoulli bandits, choice outcomes are drawn from an arm-specific Bernoulli distribution. Bernoulli bandits together with Gaussian bandits are the most commonly studied variant of multi armed bandits, both in theoretical and applied machine learning \cite{chapelle2011empirical,lu2019adaptive,liu2018change,kaufmann2012thompson} and experimental cognitive neuroscience \cite{wilson2012inferring,steyvers2009bayesian,behrens2007learning}. This is followed by an introduction of three algorithms: we start with the derivation of the learning rules based on variational SMiLE, and introduce different action selection algorithms. Importantly, in active inference we will derive an approximate action selection scheme comparable in form to the well known UCB algorithm. Finally, we empirically evaluate the performance of different algorithms, and discuss the implications of the results for the fields of machine learning and cognitive neuroscience. 

% --------------------------------------------------------------------------------------------------
% Tasks
% --------------------------------------------------------------------------------------------------

\section{The multi-armed bandit problem}

The bandit problem is a sequential game between an agent and an environment \cite{lattimore2020bandit}. The game is played in a fixed number of rounds (a horizon), where in each round the agent chooses an action (commonly referred to as a bandit arm). In response to the action, the environment delivers an outcome (e.g. a reward, punishment, or null). The goal of the agent is to develop a policy that allocates choices so as to maximise cumulative reward over all rounds. Here, we will be concerned with a bandit problem where the agent chooses between multiple arms (actions), a so-called multi-armed bandit (MAB). A well-studied canonical example is the stochastic stationary bandit, where rewards are drawn from arm-specific and fixed (stationary) probability distributions \cite{slivkins2019introduction}. 

Here, the exploration-exploitation trade-off stems from the uncertainty of the agent about how the environment is delivering the rewards, and from the fact that the agent observes outcomes only for the chosen arms, that is, it has only incomplete information about the environment. Hence, the agent obtains not only rewards from outcomes but also learns about the environment by observing the relation between an action and its outcome. Naturally, more information can be obtained from arms that have been tried fewer times, thus creating a dilemma between obtaining information, about an unknown reward probability of an arm, or trying to obtain a reward from a familiar arm. Importantly, in bandit problems there is no need to plan ahead because available choices and rewards in the next run are not affected by current choices \footnote{Note that this type of dependence between current and future choice sets, or rewards, would convert the bandit problem into a reinforcement learning problem. It makes the exploration-exploitation trade-off more complex and optimal solutions cannot be derived beyond trivial problems.}. The lack of need for planning simplifies the problem substantially and puts a focus on the exploration-exploitation trade-off, making the bandit problem a standard test-bed for any algorithm that purports to address the trade-off \cite{mattos2019multi}.

Bandit problems were theoretically developed largely in statistics and machine learning, usually focusing on the canonical stationary bandit problem \cite{lattimore2020bandit,lai1985asymptotically,slivkins2019introduction,auer2002finite,kaufmann2012bayesian,kaufmann2012thompson}. However, they also play an important role in cognitive neuroscience and psychology, where they have been applied in a wide range of experimental paradigms, investigating human learning and decision-making rather than optimal performance. Here dynamic or non-stationary variants have been used more often, as relevant changes in choice-reward contingencies in everyday environments of humans and other animals are typically hidden and stochastic \cite{schulz2019algorithmic, gottlieb2013information, wilson2012inferring, cohen2007should, behrens2007learning}. We focus on a switching bandit, a particularly popular variant of the dynamic bandit where contingencies change periodically and stay fixed for some time between switches \cite{izquierdo2017neural,markovic2019predicting,iglesias2013hierarchical,wilson2012inferring,racey2011pigeon,behrens2007learning}. The canonical stationary bandit has been influential in cognitive neuroscience and psychology as well  \cite{steyvers2009bayesian,stojic2020uncertainty,wilson2014humans,reverdy2014modeling}, in particular when combined with side information or context to investigate structure or function learning \cite{acuna2008bayesian,stojic2020s,schulz2018putting,schulz2020finding}. 

Note that many experimental tasks, even if not explicitly referred to as bandit problems, can be in fact reformulated as an equivalent bandit problem. The often used reversal learning task \cite{izquierdo2017neural}, for example, corresponds to a dynamic switching two-armed bandit \cite{clark2004neuropsychology}, and the popular go/no-go task can be expressed as a four-armed stationary bandit \cite{guitart2011action}, as another example. Furthermore, various variants of the well-established multi-stage task \cite{daw2011model} can be mapped to a multi-armed bandit problem, where the choice of arm corresponds to a specific sequence of choices in the task \cite{dezfouli2012habits}. 

In summary, we will perform a comparative analysis on two types of bandit problems: stationary stochastic and switching bandit. In this section, we first describe each bandit problem formally in an abstract way and then specify the particular instantiations we use in our computational experiments.

\subsection{Stationary stochastic bandit}
\label{sec:ssb}

A stationary stochastic bandit with finitely many arms is defined as follows: in each round $t \in \cbr{1, \ldots, T}$ the agent chooses an arm or action $k$ from a finite set of $K$ arms, and the environment then reveals an outcome $o_{t}$ (e.g. reward or punishment). The stochasticity of the bandit implies that outcomes $o_{t}$ are {\it i.i.d}. random variables drawn from a probability distribution  $o_{t} \sim p(o_t|\vec{\theta}, a_t)$. In Bernoulli bandits, these are draws specifically from a Bernoulli distribution for which outcomes are binary, that is, $o_{t} \in \cbr{0, 1}$, where each arm $k$ has a reward probability $\theta_{k}$ that parametrises the Bernoulli distribution. Hence, we can express the observation likelihood as
\begin{equation}
    p\del{o_t | \vec{\theta}, a_t=k} = \theta_{k}^{o_{t}}\left( 1 - \theta_{k} \right)^{1 - o_{t}}
\end{equation}
where $a_t$ denotes the chosen arm on trial $t$. In stationary bandits reward probabilities of individual arms $\theta_{k}$ are fixed for all trials $t$. We use $k^*$ to denote an optimal arm associated with the maximal expected reward $\theta_{k^*}$. 

In our computational experiments we follow a setup that has been used in previous investigations of stationary stochastic bandits \cite{chapelle2011empirical}: We consider the variant of the problem in which all but the best arm $k^*$ have the same reward probability $\theta_k = p = \frac{1}{2}, \forall k \in \{1,\ldots, K\} \setminus  \{k^*\}$. The probability of the best arm is set to $\theta_{k^*} = p + \epsilon$, where $0 < \epsilon < \frac{1}{2}$. The number of arms $K$ and the mean outcome difference $\epsilon$ modulate the task difficulty. The more arms and the lower the reliability, the more difficult is the problem. To understand how task difficulty influences the performance of different action selection algorithms, in the experiments we systematically vary $K \in \{10,20,40,80\}$ and $\epsilon \in \{0.05,0.10,0.20\}$ steps. Note that the larger number of arms ($K>10$) is a standard setting in machine learning benchmarks, as many industrial applications of multi-armed bandits contain a large number of options \cite{slivkins2019introduction}. In contrast, in experimental cognitive neuroscience one typically considers only a small number of options (e.g. two or three), to reduce the task complexity, thus, the training time and the experiment duration. {Interestingly, when humans are exposed to a large number of options it appears that they are a priori discounting a number of options, thus simplifying the tasks for themselves. The exact neuronal and computational mechanisms of option discounting in complex problems are still a topic of extensive research \cite{tversky1972elimination,reutskaja2011search,lieder2017strategy,stojic2020uncertainty} and go beyond the the scope of this paper.}

\subsection{Switching bandit}
\label{sec:switching}

A switching bandit is a dynamic multi-armed bandit, which, as the stationary bandit, is characterised by a set of $K$ arms, where each arm $k \in \cbr{1, \ldots, K}$ is associated with an {\it i.i.d}. random variable $o_{t}$ at a given time step $t \in \cbr{1, \ldots, T}$. However, in contrast to the stationary bandit problem, outcomes are drawn from a time-dependent Bernoulli probability distribution 
\begin{equation}
    p(o_t|\vec{\theta}_t, a_t = k)=  \theta_{t, k}^{o_t} \del{1 - \theta_{t,k}}^{1-o_t}.
\end{equation}

We use $k_t^*$ to denote the optimal arm associated with the maximal expected reward $\theta_{t, k^*_t}$ at trial $t$; hence, $k_t^* = \argmax_k \theta_{t, k}$. 

In the switching bandit \cite{cheung2019hedging, besson2019generalized} the reward probability $\theta_{t, k}$ changes suddenly but is otherwise constant. Here we use the same reward probability structure as in the stationary bandits, but change the optimal arm $k_t^*$ with probability $\rho$ as follows
\begin{equation}
\begin{split}
    j &\sim p(j_t) = \rho^{j_t} \del{1 - \rho}^{1 - j_t} \\
    k_t^* &\sim \left\{ \begin{array}{cc}
        \delta_{k^*_{t-1}, k_t^*}, & \textrm{ if } j_t = 0,  \\
        \frac{1 - \delta_{k^*_{t-1}, k_t^*}}{K - 1},  & \textrm{ if } j_t = 1,
    \end{array}
    \right.
\end{split}
\end{equation}
where $\delta_{i,j}$ denotes the Kronecker delta, and $j_t$ denotes an auxiliary Bernoulli random variable representing the presence or absence of a switch on trial $t$. The optimal arm is always associated with the same reward probability $\theta_{t, k_t^*} = p + \epsilon$ and the probability of all other arms is set to the same value $\theta_{t, k} = p =\frac{1}{2}, \forall k \neq k_t^*$. In the experiments with the switching bandit problem we systematically vary {$K \in \{10,20,40,80\}$ and $\rho \in \{0.005,0.01,0.02,0.04\}$, and $\epsilon \in \{0.05, 0.10, 0.20\}$ sets}.

In addition, we will consider the possibility that the task difficulty changes over time. Specifically, we will consider a setup in which the mean outcome difference $\epsilon$ is not fixed, and changes over time. We obtain an effective non-stationary $\epsilon$ by introducing a time evolution of the reward probabilities $\theta_{t, k}$. At each switch ($j_t=1$) point, we generate the reward probabilities from a uniform distribution. Hence, the dynamics of the switching bandit with non-stationary difficulty can be expressed with the following transition probabilities
\begin{equation}
\begin{split}
       p(\theta_{t, k}| \theta_{t-1,k}, j_t) &= \left\{ \begin{array}{cc}
        \delta( \theta_{t,k} - \theta_{t-1,k}), & \textrm{for } j_t = 0,  \\
        \mathcal{B}e\del{1, 1},  & \textrm{for } j_t=1,
    \end{array}
    \right.
\end{split}
\end{equation}
where $\delta(x)$ denotes Dirac's delta function, and $\mathcal{B}e\del{1, 1}$ a uniform distribution on $\sbr{0, 1}$ interval, expressed as the special case of a Beta distribution.

\subsection{Evaluating performance in bandit problems}

A standard approach to evaluate the performance of different decision making algorithms in bandit problems is regret analysis \cite{lattimore2020bandit,blum2007learning}, and we will therefore use it here as a primary measure. Regret is typically defined as an external measure of performance which computes a cumulative expected loss of an algorithm relative to an oracle which knows the ground truth and always selects the optimal arm $k*$. If we define the cumulative expected reward of an agent, up to trial $T$, that chose arm $a_t$ on trial $t$ as $\sum_{t=1}^T \theta_{t, a_t}$ then the (external) cumulative regret is defined as 
\begin{equation}
R_T = T \theta_{t, k^*} - \sum_{t=1}^T \theta_{t,a_t}.
\end{equation}

The cumulative regret can also be viewed as a retrospective loss, which an agent playing the bandit game can estimate after it learns which arm was optimal. This definition makes sense for stationary stochastic bandits and in the limit of $T \rightarrow \infty$. In practice, the cumulative regret $R_T$ of a specific agent playing the game will be a function of the sequence of observed outcomes $o_{1:T}$, the sequence of chosen arms $a_{1:T}$, and a selection strategy of the given agent. 

We additionally introduce a regret rate measure, a time average of the cumulative regret
\begin{equation}
  \tilde{R}_T = \frac{1}{T} R_T = \theta_{k^*} - \frac{1}{T}\sum_{t=1}^T \theta_{t, a_t}.
  \label{eq:reg_rate}
\end{equation}

In the case of stationary bandits a decision making algorithm is considered consistent if $\lim_{T\rightarrow\infty}\tilde{R}_T=0$ and asymptotically efficient if its cumulative regret approaches the following lower bound as $T \rightarrow \infty$ \cite{lai1985asymptotically}
\begin{equation}
\begin{split}
   \tilde{R}_T &\geq \underline{R}_T \\
   \underline{R}_T &= \ln(T) \sum_{i\neq k^*} \frac{\theta_{k^*} - \theta_i}{D_{KL}\del{p_{\vec{\theta}}(o_t|i)||p_{\vec{\theta}}(o_t| k^*)}} + const. \equiv \omega(K, \epsilon) \ln T + const.
\end{split}
\end{equation}

In our case of Bernoulli bandits and specifically structured reward probabilities (see \nameref{sec:ssb} subsection) the Kullback-Leibler divergence between outcome likelihoods of any arm $i\neq k^*$ and the arm $k^*$ associated with highest reward probability becomes
\begin{equation}
   D_{KL}\del{p_{\vec{\theta}}(o_t|k^*)||p_{\vec{\theta}}(o_t| i)} = -\frac{1}{2} \ln\del{1 - 4\epsilon^2} \approx 2 \epsilon^2.
\end{equation}

Hence, the lower bound to the cumulative regret becomes approximately
\begin{equation}
\label{eq:limit}
    \underline{R}_T = 2 \epsilon \frac{K-1}{\ln \del{1 + 4\epsilon^2}} \ln T \approx  \frac{K-1}{2 \epsilon} \ln T.
\end{equation}

In addition, we can define an upper bound in terms of a random choice algorithm, which selects any arm with same probability on every trial. In the case of random and uniform action selection the cumulative regret becomes
\begin{equation}
 \bar{R}_T = T \epsilon \frac{K - 1}{K}   
\end{equation}

Note that the cumulative regret is an external quantity not accessible to an agent, which has uncertain beliefs about the reward probabilities of different arms. Although, in stationary bandits the cumulative regret can reveal how efficient an algorithm is in accumulating reward in the long term, it tells us little about how efficient an algorithm is in reducing regret in the short-term. This short-term efficiency is particularly important for dynamic bandits as an agent has to switch constantly between exploration and exploitation. Therefore, to investigate short-term efficiency of the algorithm, specifically in the dynamic context, we will analyse the regret rate, instead of the commonly used cumulative regret (see \cite{raj2017taming}).

% --------------------------------------------------------------------------------------------------
% Algorithms
% --------------------------------------------------------------------------------------------------

\section{Algorithms}

Bandit algorithms can be thought of as consisting of two parts: (i) a learning rule that estimates action values, and (ii) an action-selection strategy that uses the estimates to choose actions and effectively balance between exploration and exploitation. As described in the previous section, for the canonical stationary problem a good bandit algorithm achieves a regret that scales sub-linearly with the number of rounds (see \cref{eq:limit}). Intuitively, this means that the algorithm should be reducing exploration and allocating more choices over time to arms with high expected value. The relevant question is how to reduce exploration concretely? Naturally, this is a fine balancing act: reducing exploration too quickly would potentially result in false beliefs about the best arm, hence repeatedly choosing sub-optimal arms and accumulating regret. In contrast, reducing exploration too slowly would result in wasting too many rounds exploring sub-optimal arms and again accumulating regret. 

For comparison with the algorithm based on active inference, we focus on two popular classes of bandit algorithms that are known to hit the right balance: the (Bayesian) upper confidence bound (B-UCB) \cite{auer2002finite,kaufmann2012bayesian} and (optimistic) Thompson sampling (O-TS) \cite{chapelle2011empirical,thompson1933likelihood,kaufmann2012thompson,raj2017taming} algorithms. {Our aim is not to exhaustively test bandit algorithms, but to provide a proof-of-concept and evaluate whether active inference based algorithms are viable bandit algorithms. Hence, we have to necessarily ignore a multitude of other bandit algorithms that would also be interesting competitors, but are in our judgement less popular. For example, there are other interesting information-directed algorithms for the stationary case \cite{russo2014learning,frazier2008knowledge}, or algorithms that are more finely tuned for the switching bandits \cite{garivier2011upper,besbes2014stochastic,allesiardo2017non}.} {Note that $\epsilon$-greedy or Softmax action-selection strategies \cite{sutton2018reinforcement}, frequently used in reinforcement learning, have fixed exploration, and consequently poor regret performance in open ended bandit problems (i.e. problems with an unknown time horizon). There are variants of these strategies where exploration parameters, $\epsilon$ in $\epsilon$-greedy and $\tau$ in Softmax, are reduced with specific schedules \cite{auer2002finite}. However, choosing a schedule is based on heuristics and parameters are difficult to tune. Hence, we do not include these types of strategies in our comparisons.}

In what follows we decompose active inference and {the other two bandit algorithms into two components:} the learning rule and the action selection strategy. We derive learning rules from an approximate Bayesian inference scheme and keep the rules fixed across action selection strategies, and modify only the action selection strategy. This setup allows us to have a fair comparison between active inference and the competing bandit algorithms. Finally, we will use the same action selection strategies for both stationary and dynamic bandit problem, and {derive} parameterised learning rules \chg{e}{that can} account for the presence or absence of changes.

\subsection{Shared learning rule - variational SMiLe}
\label{sec:smile}
 
To derive the belief update equations we start with a hierarchical generative model described here and apply variational inference to obtain approximate learning rules. The obtained belief update equations correspond to the variational surprise minimisation learning (SMiLe) rule \cite{liakoni2021learning,markovic2016comparative}. Importantly, we recover the learning rules for the stationary bandit (see~\ref{eq:stat_lr}) as a special case when changes are improbable.

\begin{figure}[ht!]
\centering
\includegraphics[width=0.6\textwidth]{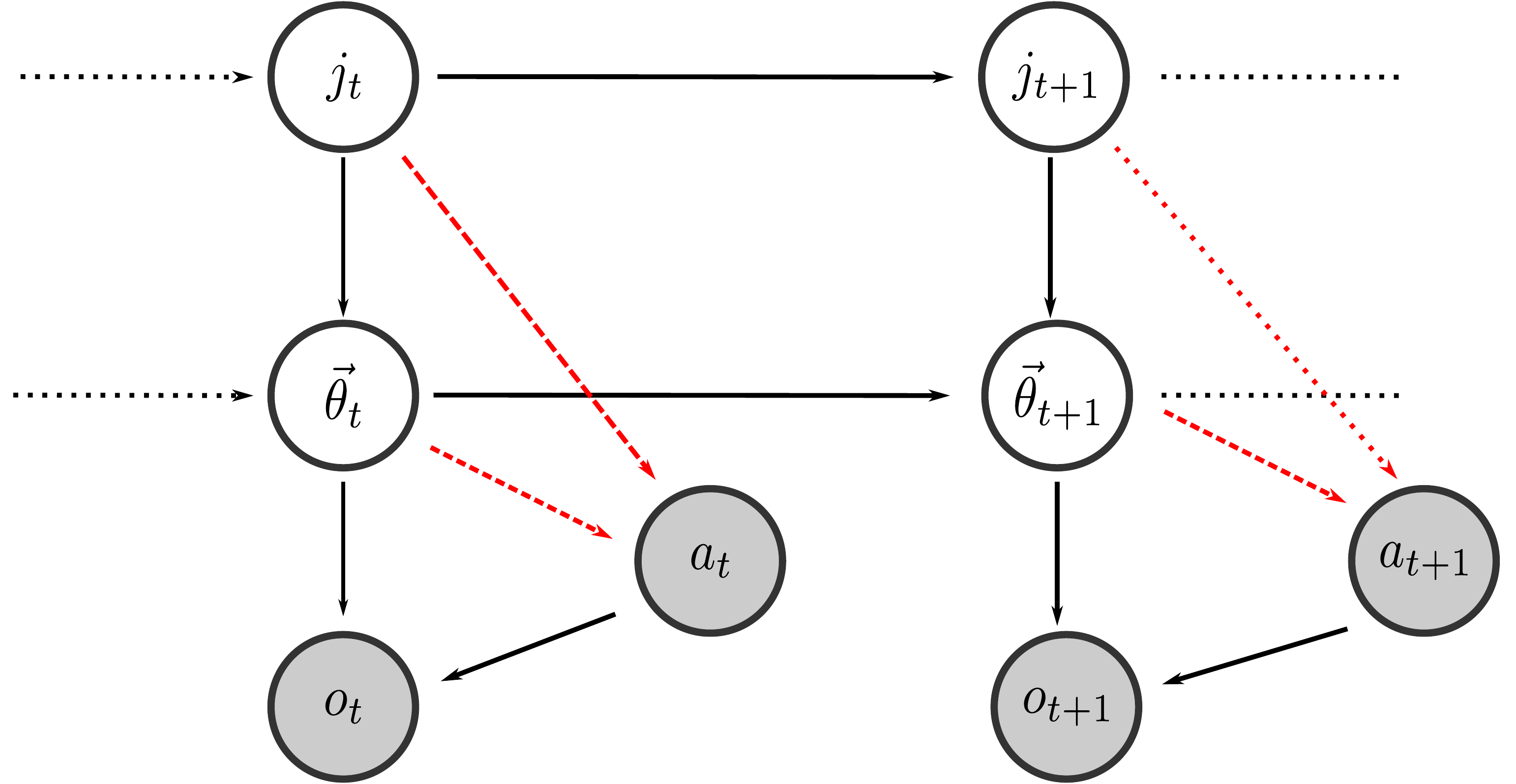}
\caption{{\bf Graphical representation of the generative model}. Shaded circles denote observables and transparent circles denote latent random variables. Note that unlike the outcomes $o_t$, which depend on latent states $\vec{\theta}_t$ actions are generated from beliefs (probability distribution) $p(\vec{\theta}_t, j_t)$ about current reward probabilities $\vec{\theta}_t$ and change probability $j_t$. Hence, we use dashed red arrows to underline the causal dependence on beliefs, in contrast to the causal dependence on latent states marked with black arrows. In practice, beliefs about latent states are fully described with parameters of a Beta distribution $(\alpha_{k, t-1}, \beta_{k, t-1})$ associated with each arm $k$ and the explicit knowledge of the change probability $\rho$. Hence, Bayesian bandit algorithms will differ only in the way they map the beliefs into actions.}
\label{fig:gen_model}
\end{figure}

We will express the hierarchical generative model of choice outcomes $o_{1:T}=\del{o_1, \ldots, o_T}$ as the following joint distribution
\begin{equation}
    p(o_{1:T}, \vec{\theta}_{1:T}, j_{1:T}| a_{1:T}) = \prod_{t=1}^T p(o_t|\vec{\theta}_t, a_t) p(\vec{\theta}_t|\vec{\theta}_{t-1}, j_t) p(j_t),
\end{equation}
where the observation likelihood corresponds to the Bernoulli distribution. Hence,
\begin{equation}
    p(o_t|\vec{\theta}_t, a_t) = \prod_{k=1}^K \sbr{\theta_{t,k}^{o_t}\del{1-\theta_{t,k}}^{1-o_t}}^{\delta_{a_t,k}}.
\end{equation}

If no change ($j_t=0$) occurs on a given trial $t$ the reward probabilities are fixed, $\vec{\theta}_t = \vec{\theta}_{t-1}$. Otherwise, if a change occurs ($j_t=1$), a new value is generated for each arm from some prior distribution $\mathcal{B}e\del{\alpha_0,\beta_0}$. Formally, we can express this process as
\begin{equation}
    p(\vec{\theta}_t|\vec{\theta}_{t-1}, j_t) = \left\{ \begin{array}{ll} \delta\del{\vec{\theta}_t - \vec{\theta}_{t-1}}, & \textrm{ if } j_t=0 \\
    \prod_{i=1}^K \mathcal{B}e \del{\alpha_0, \beta_0} & \textrm{ if } j_t=1  \end{array} \right.
\end{equation}

Similarly, the probability that a change in reward probabilities occurs on a given trial is $\rho$, hence we \chg{f}{express the probability of change occurring on trial $t$ as the following Bernoulli distribution}
\begin{equation}
    p\del{j_t} = \rho^{j_t}\del{1 -\rho}^{1-j_t}
\end{equation}

The Bayesian approach requires us to specify a prior. The prior over reward probabilities associated with each arm $p(\vec{\theta}_0) = p(\theta_{0,1}, \ldots, \theta_{0, K})$ is given as the product of conjugate priors of the Bernoulli distribution, that is, the Beta distribution
\begin{equation}
    p(\vec{\theta}_0) = \prod_{k=1}^K \mathcal{B}e(\alpha_{0,k}, \beta_{0,k}),
\end{equation}
where we initially set the prior to a uniform distribution, $\alpha_{0,k}, \beta_{0,k} = 1, \forall \: k$. {In  \cref{fig:gen_model} we show the graphical representation of the generative model.}

Hence, given the Bayes rule at time step $t$
\begin{equation}
     p(\vec{\theta}_t, j_t|o_{1:t}, a_{1:t}) \propto p(o_t|\vec{\theta}_t, a_t) p(\vec{\theta}_t, j_t|o_{1:t-1}),
\end{equation}
we can express the exact marginal posterior beliefs over reward probabilities $\vec{\theta}_t$ as
\begin{equation}
\begin{split}
   p\del{\vec{\theta}_t| o_{1:t}, a_{1:t}} &= \del{1 - \gamma_t} p\del{\vec{\theta}_t|j_t=0, o_{1:t}, a_{1:t}} \\ 
   &+ \gamma_t p\del{\vec{\theta}_t|j_t=1, o_t, a_t}
\end{split}
\label{eq:exact_marg}
\end{equation}
where and $a_{1:t}$ corresponds to the sequence of chosen arms, and $\gamma_t$ corresponds to the marginal posterior probability that a change occurred on trial $t$. \chg{g}{We obtain the posterior change probability as follows}
\begin{equation}
\begin{split}
    \gamma_t  &= \gamma\del{S_{BF}^t, m} \\
    \gamma(S, m) &= \frac{mS}{1 + mS} \\
    S_{BF}^t &= \frac{p\del{o_t|j_t=0, a_t, o_{t:t-1}}}{p\del{o_t|j_t=1, a_t, o_{1:t-1}}} \\
    m & = \frac{\rho}{1 - \rho}
\end{split}
\label{eq:change_prob}
\end{equation}

{The exact marginal posterior in \cref{eq:exact_marg} will not belong to the Beta distribution family, making the exact inference analytically intractable, as each iteration of the belief update results in a novel distribution family with ever increasing complexity. In practice, there are numerous ways one can perform approximate inference in dynamic Bernoulli bandits \cite{adams2007bayesian,mellor2013thompson,lu2019adaptive,alami2020restarted}. Here we will focus on the method based on variational inference due to its simplicity and efficiency. Although, more optimal inference methods do exist, we do not expect them to change the relative performance of different decision algorithms as for Bayesian bandits we can always use the same (most optimal) learning rule for all Bayesian decision algorithms.} 

To obtain the learning rule we constrain the joint posterior to an approximate, fully factorised, form, expressed as
\begin{equation}
    p(\vec{\theta}_t, j_t| o_{1:t}, a_{1:t}) \approx Q(j_t)\prod_{k=1}^K Q(\theta^k_t).
\end{equation}
Applying the variational calculus results in the following variational SMiLe rule (for more details on derivations of the SMiLe rule see \cite{liakoni2021learning})
\begin{equation}
\begin{split}
    \alpha_{t, k} &= (1 - \gamma_t)\alpha_{t-1, k} + \gamma_t\alpha_0 + \delta_{a_t, k} o_t\\
    \beta_{t, k} &= (1 - \gamma_t)\alpha_{t-1, k} + \gamma_t\alpha_0 + \delta_{a_t, k} \del{1 - o_t}
\end{split}
\end{equation}
for the parameters of the Beta distributed factors $Q(\theta_{t,k}) = \mathcal{B}e\del{\alpha_{t,k}, \beta_{t,k}}$. The categorically distributed change probability $Q(j_t=1) = \gamma_t$ is update based on \cref{eq:change_prob}.

Note that for a stationary environment the changes are improbable, hence $\rho=0$ and consequently $\gamma_t = 0$ for every $t$. This implies that for the stationary bandit we recover the following learning rules
\begin{equation}
    \begin{split}
    \alpha_{t, k} &= \alpha_{t-1, k} + o_t \cdot \delta_{a_t, k}, \\
    \beta_{t, k} &= \beta_{t-1,k} + (1-o_t) \delta_{a_t, k},
    \end{split}
\label{eq:stat_lr}
\end{equation}
that correspond to the exact Bayesian inference over the stationary Bernoulli bandit problem, as in absence of changes the Beta prior corresponds to a conjugate prior of a Bernoulli likelihood.

\subsection{Action selection}

\subsubsection{Active inference}

One view on the exploration-exploitation trade-off is that it can be formulated as an uncertainty-reduction problem \cite{schwartenbeck2019computational}, where choices aim to resolve expected and unexpected uncertainty about hidden properties of the environment \cite{soltani2019adaptive}. This leads to casting choice behaviour and planning as a probabilistic inference problem \cite{kaplan2017navig, friston2017temporal, friston2017process, mirza2016scene, friston2016learning}, as expressed by active inference. Using this approach, different types of exploitative and exploratory behaviour naturally emerge \cite{schwartenbeck2019computational}. In active inference, decision strategies (behavioural policies) are chosen based on a single optimisation principle: minimising expected surprisal about observed and future outcomes, that is, the expected free energy \cite{friston2017process}. Formally, we express the expected free energy of a choice $a$ on trial $t$ as
\begin{equation}
    \begin{split}
    G_t(a) & =  \underbrace{D_{KL}\del{Q(o_t |a_t = a)||P(o_t)}}_{\textrm{Risk}} + \underbrace{E_{Q(\vec{\theta})}\left[H\sbr{o_t|\vec{\theta}, a_t=a} \right]}_{\textrm{Ambiguity}} \\
    & = - \underbrace{E_{Q(o_t|a_t=a)}\left[ \ln P(o_t)\right]}_{\textrm{Extrinsic value}} \\ &\quad\: - \underbrace{E_{Q(o_t|a_t=a)}\left[ D_{KL}\del{ Q\del{\vec{\theta}, j_t|o_t, a_t=a}|| Q\del{\vec{\theta}, j_t}} \right]}_{\textrm{Intrinsic value/Novelty}}
    \end{split}
\label{eq:efe}
\end{equation}
where $Q\del{\vec{\theta}, j_t} = p\del{\vec{\theta}, j_t|o_{1:t-1}}$,  $Q(o_t|a_t) = \int \dif \vec{\theta} p(o_t|\vec{\theta}, a_t) Q\del{\vec{\theta}} $, $P(o_t)$ denotes prior preferences over outcomes, $H\sbr{o_t|\vec{\theta}, a_t}$ denotes the conditional entropy of observation likelihood $p\del{o_t|\vec{\theta}, a_t}$, and $D_{KL}(p||q)$, stands for the Kullback-Leibler divergence. Then, a choice $a_t$ is made by selecting the action with the smallest expected free energy\footnote{In usual applications of active inference for understanding human behaviour, rather than minimising the expected free energy one would sample actions from posterior beliefs about actions (cf. planning as inference \cite{botvinick2012planning,attias2003planning}). This becomes useful when fitting empirical choice behaviour in behavioural experiments \cite{markovic2019predicting,schw2016}.}
\begin{equation}
\label{eq:exact_ai}
    a_t = \argmin_a G_t(a),
\end{equation}
where we consider the simplest form of active inference, as in other bandit algorithms, one-step-ahead beliefs about actions.

Note that in active inference, the most likely action has dual imperatives, implicit within the expected free energy acting as the loss function (see the different decomposition in \cref{eq:efe}): The expected free energy can, on one hand, be decomposed into ambiguity and risk. On the other hand, it can be understood as a combination of intrinsic and extrinsic value, where intrinsic value corresponds to the expected information gain, and the extrinsic value to the expected value. The implicit information gain or uncertainty reduction pertains to beliefs about the parameters of the likelihood mapping, which has been construed as novelty \cite{kaplan2018planning,schwartenbeck2013exploration}. \chg{efe}{Therefore, selecting actions that minimise the expected free energy dissolves the exploration-exploitation trade-off, as every selected action tries to maximise the  expected value and the expected information gain at the same time.}

To express the expected free energy, $G_t(a)$, in terms of beliefs about arm-specific reward probabilities, we will first constrain the prior preference to the following Bernoulli distribution
\begin{equation}
    P(o_t) = \frac{1}{Z(\lambda)} e^{o_t \lambda} e^{-(1-o_t) \lambda}.
    \label{eq:preference}
\end{equation}

In active inference, prior preferences determine whether a particular outcome is attractive or rewarding. Here we assume that agents prefer outcome $o_t=1$ over outcome $o_t=0$. Hence, we specify payoffs or rewards with prior preferences over outcomes that have an associated precision $\lambda$, where $\lambda \geq 0$. The precision parameter $\lambda$ determines the balance between epistemic and pragmatic imperatives. When prior preferences are very precise, corresponding to large $\lambda$, the agent becomes risk sensitive and will tend to forgo exploration if the risk ({\it i.e.}, the divergence between predicted and preferred outcomes, see Eq.~\ref{eq:efe}) is high. Conversely, a low lambda corresponds to an agent which is less sensitive to risk and will engage in exploratory, epistemic behaviour, until it has familiarised itself with the environment ({\it i.e.}, the latent reward probabilities \chg{h}{of different arms}).

Given the following expressions for the marginal predictive likelihood, obtained as,
\begin{equation}
\begin{split}
    Q\del{o_t|a_t} &= \int \dif \vec{\theta}_t p\del{o_t|\vec{\theta}_t, a_t} Q\del{\vec{\theta}_t} = \prod_{k=1}^K \sbr{\sbr{\tilde{\mu}_{t, k}}^{o_t}\sbr{1 - \tilde{\mu}_{t, k}}^{1-o_t}}^{\delta_{a_t, k}}\\
    Q\del{\vec{\theta}} &= p\del{\vec{\theta}_t|o_{t-1:1}} = (1-\rho)\prod_{k=1}^K\mathcal{B}e\del{\alpha_{t-1, k}, \beta_{t-1, k}} + \rho \prod_{k=1}^K\mathcal{B}e\del{\alpha_{0}, \beta_{0}}\\
    \tilde{\mu}_{t, k} &= \mu_{t-1, k} + \rho \del{\frac{1}{2} - \mu_{t-1, k}} \\
    \mu_{t-1, k} &= \frac{\alpha_{t-1, k}}{\nu_{t-1, k}} \\
    \nu_{t-1, k} &= \alpha_{t-1, k} + \beta_{t-1, k}
\end{split}
\label{eq:marginal}
\end{equation}
we get the following expressions for the expected free energy
\begin{equation}
    \begin{split}
    G_t(a) & = - 2 \lambda (1-\rho)\mu_{t-1, a} + \tilde{\mu}_{t, a} \ln \tilde{\mu}_{t, a} + (1-\tilde{\mu}_{t, a}) \ln ( 1- \tilde{\mu}_{t, a}) \\
    &- (1-\rho) \sbr{\mu_{t-1, a} \psi\del{\alpha_{t-1,a}} + \del{1 - \mu_{t-1,a}} \psi\del{\beta_{t-1, a}}} \\
    &+ (1 - \rho) \sbr{\psi\del{\nu_{t-1,a}} - \frac{1}{\nu_{t-1,a}}} + const.
    \end{split}
    \label{eq:efe_est}
\end{equation}
More details on how we derive Eq.~\ref{eq:efe_est} is available in the \nameref{sec:appendix} \cref{sec:derive_efe}.

If we approximate digamma function as $\psi(x) \approx \ln x - \frac{1}{2x}$ \chg{i}{(which is valid for $x\gg 1$)}, and note that for all relevant use cases $\rho \ll 1$; then by substituting the approximate digamma expression into Eq.~(\ref{eq:efe_est}) we get the following action selection algorithm
\begin{equation}
\label{eq:approx_ai}
a_t = \argmax_{k} \sbr{ 2 \lambda \mu_{t-1, k} + \frac{1}{2\nu_{t-1, k}}}.
\end{equation}
More details on how we arrive at Eq.~\ref{eq:efe_est} is available in the \nameref{sec:appendix} \cref{sec:derive_app}.

Note that a similar exploration bonus -- inversely proportional to the number of observations -- was proposed in the context of Bayesian reinforcement learning \cite{kolter2009near} when working with Dirichlet prior and posterior distributions.

We will denote active inference agents which make choices based on the approximate expected free energy, \cref{eq:approx_ai}, with A-AI, and agents which minimise directly the exact expected free energy, \cref{eq:exact_ai}, with G-AI.

\subsubsection{Bayesian upper confidence bound}

The upper confidence bound (UCB) is a classical action selection strategy for resolving the exploration-exploitation dilemma \cite{auer2002finite}. \chg{ucb-bernoulli}{When fine-tuned for Bernoulli bandits, the action selection strategy can be defined as
\begin{equation}
    a_t = \left\{ \begin{array}{cc}
        \argmax_k \left( m_{t, k} + \frac{\ln t}{n_{t,k}} + \sqrt{  \frac{ m_{t,k} \ln t}{ n_{t, k}}} \right) & \textrm{for } t>K \\
        t & \textrm{otherwise}
    \end{array},
    \right.
\end{equation}
where $m_{t, k}$ is the expected reward of $k$-th arm and $n_{t,k}$ the number of times the $k$-th arm was selected (see \cite{chapelle2011empirical} for more details).}

However, we consider a more recent variant called Bayesian UCB \cite{kaufmann2012bayesian}, grounded in Bayesian bandits. In Bayesian UCB the best arm is selected as the one with the highest $z$-th percentile of posterior beliefs, where the percentile increases over time as $z_t = 1 - \frac{1}{t}$. Hence, we can express the action selection rule as 
\begin{equation}
\label{eq:BUCB}
    a_t = \argmax_k CDF^{-1}(z_t, \bar{\alpha}_t^k, \bar{\beta}_t^k)
\end{equation}
where $CDF(\cdot)$ denotes cumulative distribution function of Beta distributed posterior beliefs, and the parameters ($\bar{\alpha}_t^k$, $\bar{\beta}_t^k$) denote approximate sufficient statistics of the Beta distributed prior beliefs on trial $t$. Note that the exact predictive prior on trial $t$ corresponds to a mixture of two Beta distributions
\begin{equation}
p\del{\theta_t^k|o_{t-1:1}} = (1 - \rho) \mathcal{B}e\del{\alpha_{t-1}^k, \beta_{t-1}^k} + 
\rho \mathcal{B}e\del{\alpha_0, \beta_0}.
\end{equation}

As the inverse of a cumulative distribution function of the above mixture distribution is analytically intractable we will assume the following approximation
\begin{equation}
\begin{split}
    p\del{\theta_t^k|o_{1:t-1}} & \approx \mathcal{B}e\del{\bar{\alpha}_t^k, \bar{\beta}_t^k}\\
    \bar{\alpha}_t^k & = (1 - \rho) \alpha_{t-1}^k + \rho \alpha_0 \\
    \bar{\beta}_t^k & = (1 - \rho) \beta_{t-1}^k + \rho \beta_0 
\end{split}
\end{equation}

Thus, in the case of the Beta distributed prior beliefs, the inverse cumulative distribution function corresponds to the inverse incomplete regularised beta function. Hence, we can write
\begin{equation}
    CDF^{-1}(z, \alpha, \beta) = I_z^{-1}(\alpha, \beta),
\end{equation}
\chg{j}{where $I_z^{-1}(\alpha, \beta)$ corresponds to the solution of the following equation with respect to $x$
\begin{equation}
    z = \frac{\Gamma\del{\alpha + \beta}}{\Gamma\del{\alpha}\Gamma\del{\beta}} \int_{0}^x u^{\alpha - 1} \del{1- u}^{\beta - 1} \dif u ~.
\end{equation}}

\subsubsection{Thompson sampling}

Thompson sampling is traditionally associated with Bayesian bandits \cite{kandasamy2018parallelised,chapelle2011empirical,thompson1933likelihood}, where the action selection is derived from the i.i.d samples from the posterior beliefs about the reward probability. The standard algorithm corresponds to
\begin{equation}
    a_t = \argmax_k \theta^*_{t,k}, \qquad \theta^*_{t,k} \sim p\del{\theta_{t,k}|o_{1:t-1}},
\end{equation}

where $\theta^*_{t,k}$ denotes a single sample from the current beliefs about reward probabilities associated with the $k$-th arm. 

An extension of the standard algorithm, proposed in the context of dynamic bandits, is called optimistic Thompson sampling \cite{raj2017taming}, defined as 
\begin{equation}
\label{eq:OTS}
    a_t = \argmax_k \sbr{\max\del{\theta^*_{t,k}, \tilde{\mu}_{t,k}}}, \qquad \theta^*_{t,k} \sim p\del{\theta_{t,k}|o_{1:t-1}},
\end{equation}
where the expected reward probability at current trial $t$, 
$$\tilde{\mu}_{t, k} = \mu_{t-1, k} + \rho \del{\frac{1}{2} - \mu_{t-1,k}},$$ 
constrains the minimal accepted value of the sample from the prior, hence biasing the sampling towards optimistic larger values.

\chg{code-and-data-availability}{
\subsection{Code and data availability}
The code accompanying the paper is available at \href{https://github.com/dimarkov/aibandits}{github.com/dimarkov/aibandits}. The repository contains the implementation of all algorithms and scripts for execution of the simulations. The folder with jupyter notebooks contains the scripts used to generate the figures. The results of simulations, which can be used to reproduce the figures, are available at \href{https://osf.io/85ek4/}{osf.io/85ek4/}. All the simulations are controlled with a manually set seed and it should be possible to reproduce the results exactly.
}

% --------------------------------------------------------------------------------------------------
% Results
% --------------------------------------------------------------------------------------------------

\section{Results}
\label{sec:results}

In what follows, we first examine the performance of active inference based agents, A-AI (minimising approximated estimate of the expected free energy) and G-AI (minimising exact expected free energy) in the stationary Bernoulli bandits. Using the regret rate as performance criterion we analyse the dependence of agent's performance on the precision of prior preferences ($\lambda$) parameter and simultaneously verify that our approximation is good enough. After illustrating the effectiveness of A-AI (\pcref{eq:approx_ai}), in comparison to G-AI (\pcref{eq:exact_ai}), we empirically compare only the A-AI algorithm -- now in terms of the cumulative regret -- with agents using the optimistic Thompson sampling (O-TS; \pcref{eq:OTS}) and Bayesian upper confidence bound (B-UCB; \pcref{eq:BUCB}) algorithms, in the same stationary Bernoulli bandit. \chg{k}{Finally, we provide an empirical comparison of the algorithms in the case of switching Bernoulli bandit, both in scenarios with fixed and varying difficulty}.

\subsection{The stationary Bernoulli bandit}

\begin{figure}[ht!]
\centering
\includegraphics[width=\textwidth]{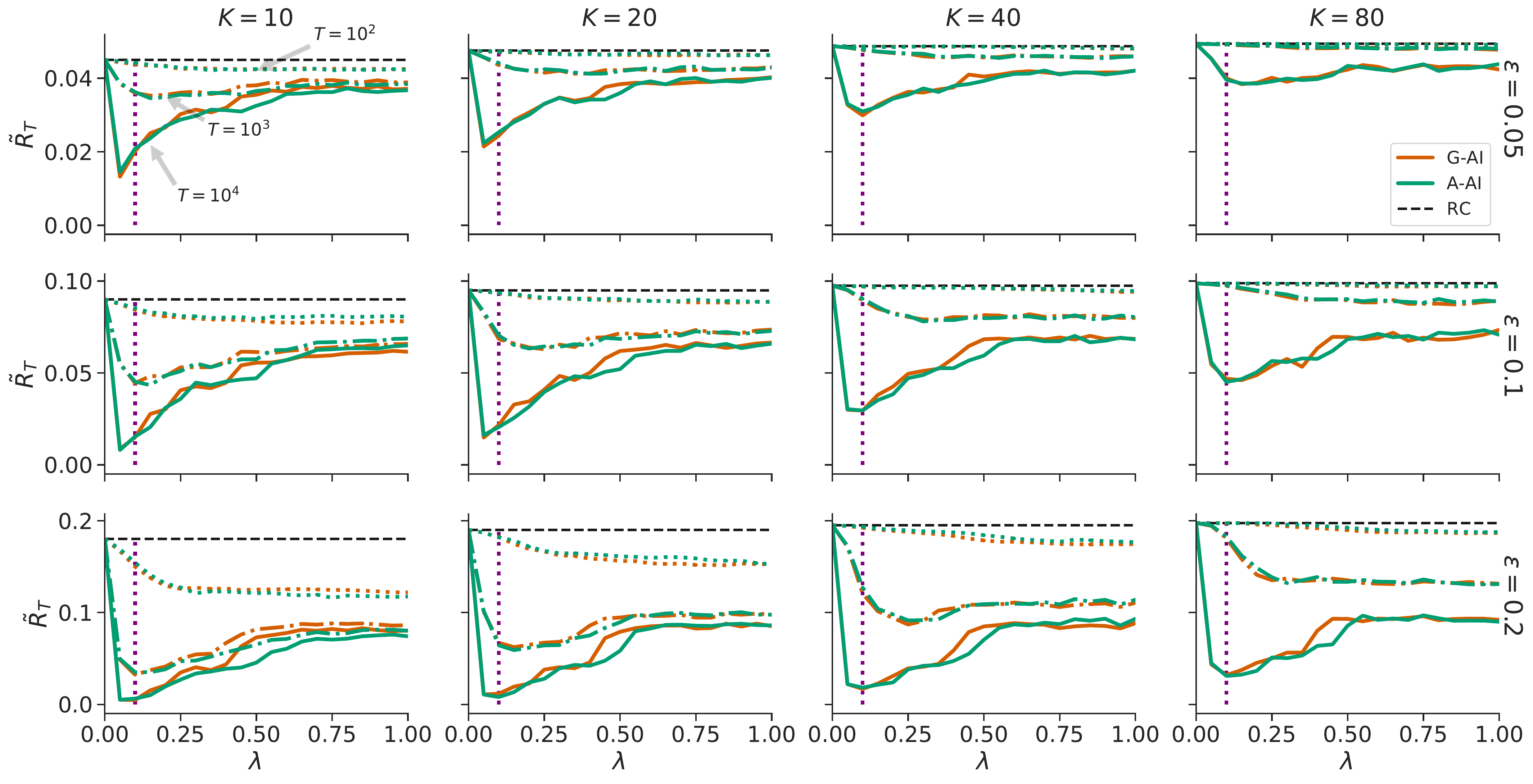}
\caption{{\bf Regret rate analysis for active inference based agents in the stationary Bernoulli bandit}. The regret rate $\tilde{R}_T$, \cref{eq:reg_rate}, for the approximate (A-AI) and the exact (G-AI) variants of active inference as a function of the precision over prior preferences $\lambda$. The coloured lines show numeric estimates obtained as an average over $N=10^3$ runs. Different line styles denote $\tilde{R}_T$ values estimated after different numbers of trials $T$: Dotted lines correspond to $T=10^2$, dotted dashed lines to $T=10^3$ and solid lines to $T=10^4$, as annotated in the top left plot. The dashed black line denotes the upper bound on the regret rate corresponding to the random (RC) agent which gains no information from the choice outcomes. The vertical doted line (purple) corresponds to $\lambda=0.1$ level, which we find to be sufficiently close to the minimum or regret rate in a range of conditions. Each column and row of the plot corresponds to different task difficulties, characterised by the number of arms $K$, and the mean outcome difference $\epsilon$, respectively.}
\label{fig:comp_ai}
\end{figure}

The precision parameter $\lambda$ acts as a balancing parameter between exploitation and exploration (\pcref{eq:approx_ai}). Hence, it is paramount to understand how $\lambda$ impacts the performance across different difficulty conditions. We expect that there will be a $\lambda^*(\epsilon, K)$ for which the active inference algorithm achieves minimal cumulative regret after a fixed number of trials $T$, for each mean outcome difference $\epsilon$ and each number of arms $K$. When the AI agent has \chg{l}{weak} preferences ($\lambda \rightarrow 0$), it would engage in exploration for longer, thereby reducing its free energy (i.e., uncertainty about the likelihood mappings), at the expense of accumulating reward. Conversely, an AI agent with \chg{m}{strong} preferences ($\lambda \rightarrow \infty$) would commit to a particular arm as soon as it had inferred that this was the arm with highest likelihood of payoffs. However, the ensuing ‘superstitious’ behaviour would prevent it from finding the best arm. To illustrate this, in \cref{fig:comp_ai} we report regret rate averages over a $N=10^3$ simulations, and compare the agents using either the approximate (A-AI) or the exact (G-AI) expected free energy for action selection. Using the regret rate simplifies the comparison, as unlike cumulative regret, the regret rate stays on the same range of values independent of trial number $T$.
\chg{differences}{Surprisingly, the A-AI algorithm achieves slightly lower regret rate in certain ranges of $\lambda$ values depending on the problem difficulty. The reason for this is that the approximate information gain used in A-AI algorithm \cref{eq:approx_ai} is always larger or equal than the exact information gain \cref{eq:efe_est}. Hence, for sufficiently low values of $\lambda$ (e.g. $\lambda < 0.25$) the behaviour is initially strongly dominated by the exploratory part of AI algorithms, and both algorithms exhibit similar regret rates. As we increase the value of $\lambda$ the exploitative part becomes more dominant in action selection. However, a higher value of $\lambda$ is required in the A-AI algorithm for the exploitative part to become dominant, as the approximate information gain is initially larger and converges slower to zero than the exact information gain. As $\lambda$ becomes sufficiently large both algorithms become equally bad (action selection start depending only on the expected value) hence the difference in regret rate disappears again. Interestingly, this performance differences are not visible in the case of switching bandits analysed later in this section.}

\chg{lambda}{Using a visual inspection of \cref{fig:comp_ai} we find the minimal regret rate -- at the asymptotic limit of large number of trials $T=10^4$, see solid lines in \cref{fig:comp_ai} -- is close to the value $\lambda=0.1$ for a range of problem difficulties}\footnote{For the hardest considered setting, corresponding to $\epsilon=0.05$, the minimum is sharp and corresponds to the value $\lambda=0.06$.}. Hence, for the \chg{n}{subsequent} between-agent comparisons we restrict the active inference agents to a fixed precision of prior preferences, $\lambda=0.1$. As both G-AI (red lines) and A-AI (green lines) achieve very similar regret rates as a function of precision $\lambda$ and number of trials $T$, we will only consider the A-AI variant for the between-agent comparison. We anticipated that even this approximate form of active inference would outperform bandit algorithms; most notably when considering short sessions in the stationary scenario: {\it i.e.}, when exploration gives way to exploitation after the agent becomes familiar with the payoffs afforded by the multi-armed options. The reason for this expectation is the exact computation of the information gain implicit within the expected free energy (see \ref{eq:efe}). 

\begin{figure}[ht!]
\centering
\includegraphics[width=\textwidth]{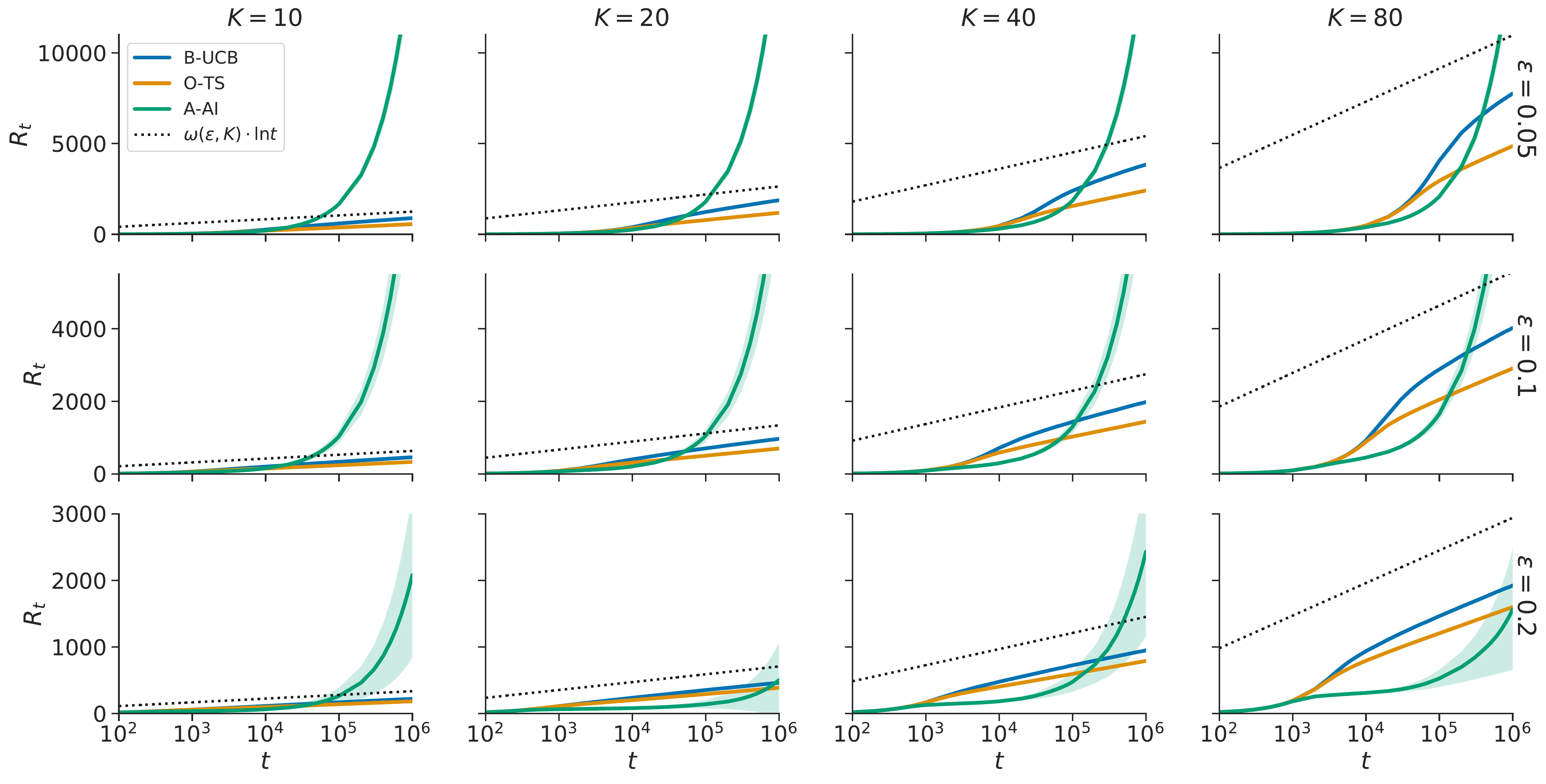}
\caption{{\bf Between-agent comparison in the stationary Bernoulli bandit.} Comparison of cumulative regret trajectories for the approximate active inference (A-AI), the optimistic Thompson sampling (O-TS), and Bayesian upper confidence bound (B-UCB) based agents. For the A-AI based agent the prior precision is set to $\lambda=0.1$, that is, to the near optimal value for a range of difficulty conditions. Solid coloured lines denote the ensemble cumulative regret average and shaded regions (not visible in every subplot) mark the 95\% confidence interval of the mean estimate. All the values are estimated as ensemble averages over $N=10^3$ simulations.}
\label{fig:cum_reg_traject}
\end{figure}

Next we compare and contrast the cumulative regret, as a function of trial number $t$, of the A-AI agents with agents based on the optimistic Thompson sampling (O-TS) and the Bayesian UCB (B-UCB) algorithms (Fig.~\ref{fig:cum_reg_traject}). The dotted lines mark the corresponding asymptotic limit (see \pcref{eq:limit}) of the corresponding problem difficulty ($\epsilon, K$). The asymptotic limit scales as $\ln t$ and defines long term behaviour of the asymptotically efficient algorithm. Note that the limit behaviour can be offset by an arbitrary constant to form a lower bound \cite{lai1985asymptotically,chapelle2011empirical}. For convenience we fix the constant to zero, and show the asymptotic curve only as a reference for long term behaviour of cumulative regret for different algorithms. 

The comparison reveals that the A-AI agent \chg{o}{on average} outperforms the bandit algorithms, but only up until some trial $t$ that depends on the task difficulty -- in the asymptotic limit the regret grows faster than logarithmic with trial number. For example, for $K=10$, A-AI outperforms bandit algorithms only up to $T=10^4$. The divergence in cumulative regret is driven by a percentage of the $N=10^{3}$ agents in the ensemble that did not not find the optimal solution and are over-confident in their estimate of the arm with the highest reward probability. \chg{histogram}{We illustrate this in \cref{fig:sup_hist} in the form of histogram of the logarithm of cumulative regret at $T=10^6$ estimated over the ensemble of $N=10^3$ agents.}  It might appear surprising, that the divergence is \chg{p}{more prominent} for the smaller number of arms. However, the reason for this is, that the smaller the number of arms is, the more chance an agent has to explore each individual arm, for a limited trial number. Hence, the agent will commit faster to a wrong arm and stay with that choice longer. Therefore, we found that our initial expectation about the performance of active inference algorithms is only partially correct. Although one could set $\lambda$ for any task difficulty in a way that active inference initially outperforms the alternative algorithms, in the asymptotic limit the high performance level will not hold. The reason for this can be seen already in \cref{fig:comp_ai}, if one notes that maximal performance (minimal regret rate) depends both on preference precision $\lambda$ and trial number $T$, for every $K, \epsilon$ tuple. 

\chg{execution-times}{We also timed the execution of all algorithms, to provide an additional measure of practicality of our approximate A-AI algorithm. All algorithms use the same learning rule, hence the only difference in execution time would come from action selection part of the algorithm. Results show that A-AI obtains the lowest time, on par with classical UCB (see \cref{tab:benchmark} in the appendix). Usual caveats with timing apply, results depend on implementation details and hardware used, as well as on the specifics of our bandit problem, hence one should be careful with generalizing from these results.
}

Although active inference based agents behave poorly in the asymptotic limit, the fact that they achieve higher performance on a short time scale suggests that in dynamic environments -- if changes occur sufficiently often -- one would get higher performance on average when compared to considered alternatives. 

\subsection{The switching bandit problem}

\begin{figure}[ht!]
\centering
\includegraphics[width=\textwidth]{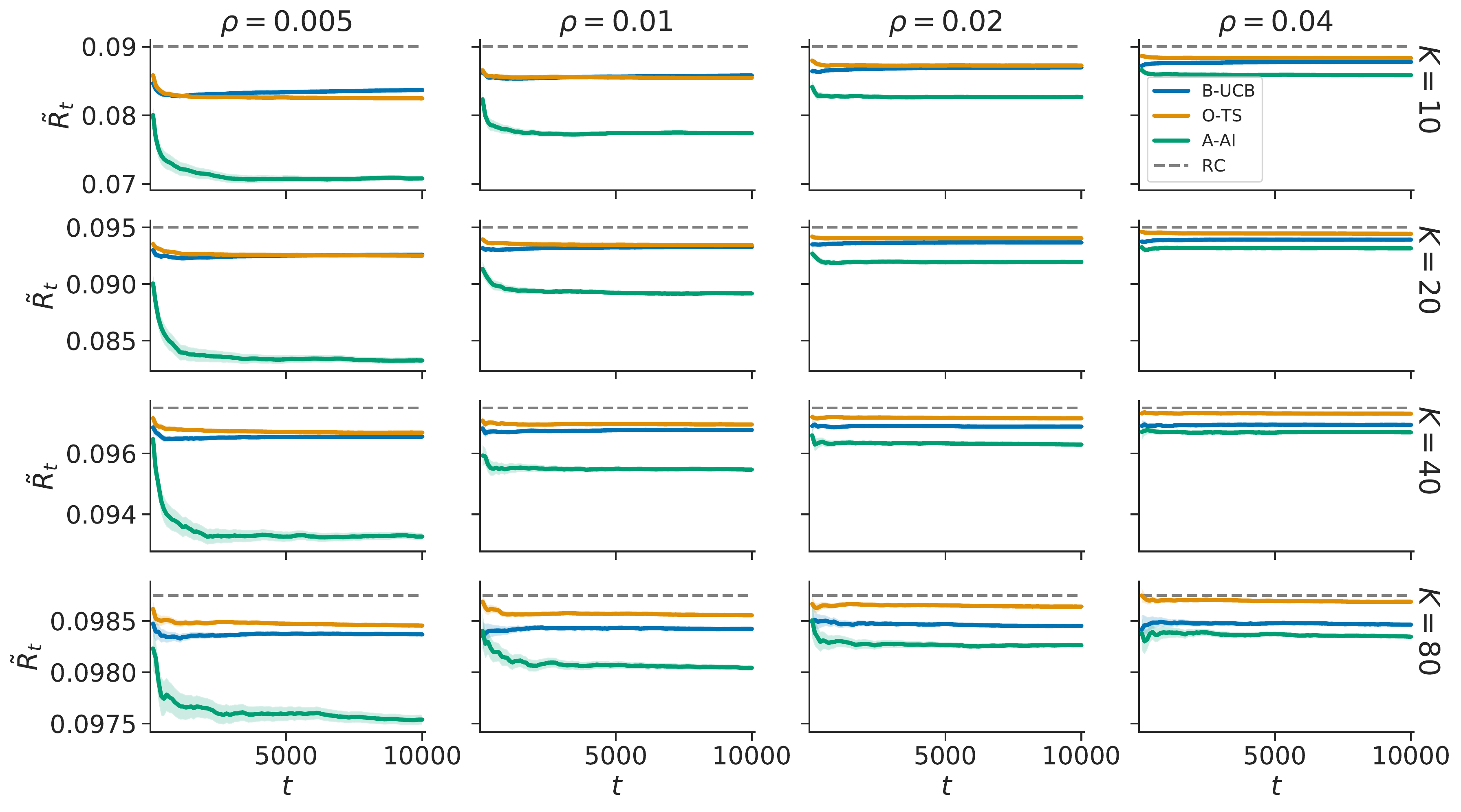}
\caption{{\bf Between-agent comparison in switching Bernoulli bandits with a fixed mean outcome difference ($\epsilon=0.1$)}. Comparison of the regret rate of approximate active inference (A-AI), optimistic Thompson sampling (O-TS), and Bayesian upper-confidence-bound (B-UCB) based agents in the switching bandit problem (see \nameref{sec:switching} subsection). Each column and row of the plot corresponds to different task difficulties, characterised by the change probability $\rho$, and the number of arms $K$. For the A-AI agents the prior precision over outcome preferences is fixed to  $\lambda=0.5$. All the values are estimated as ensemble averages over $N=10^3$ simulations, where the switching schedule is also generated randomly for each agent instance within the ensemble. The 95\% confidence intervals, although plotted, are hardly visible, implying a statistically robust comparison.}
\label{fig:dyn_comp_arms}
\end{figure}

\begin{figure}[ht!]
\centering
\includegraphics[width=\textwidth]{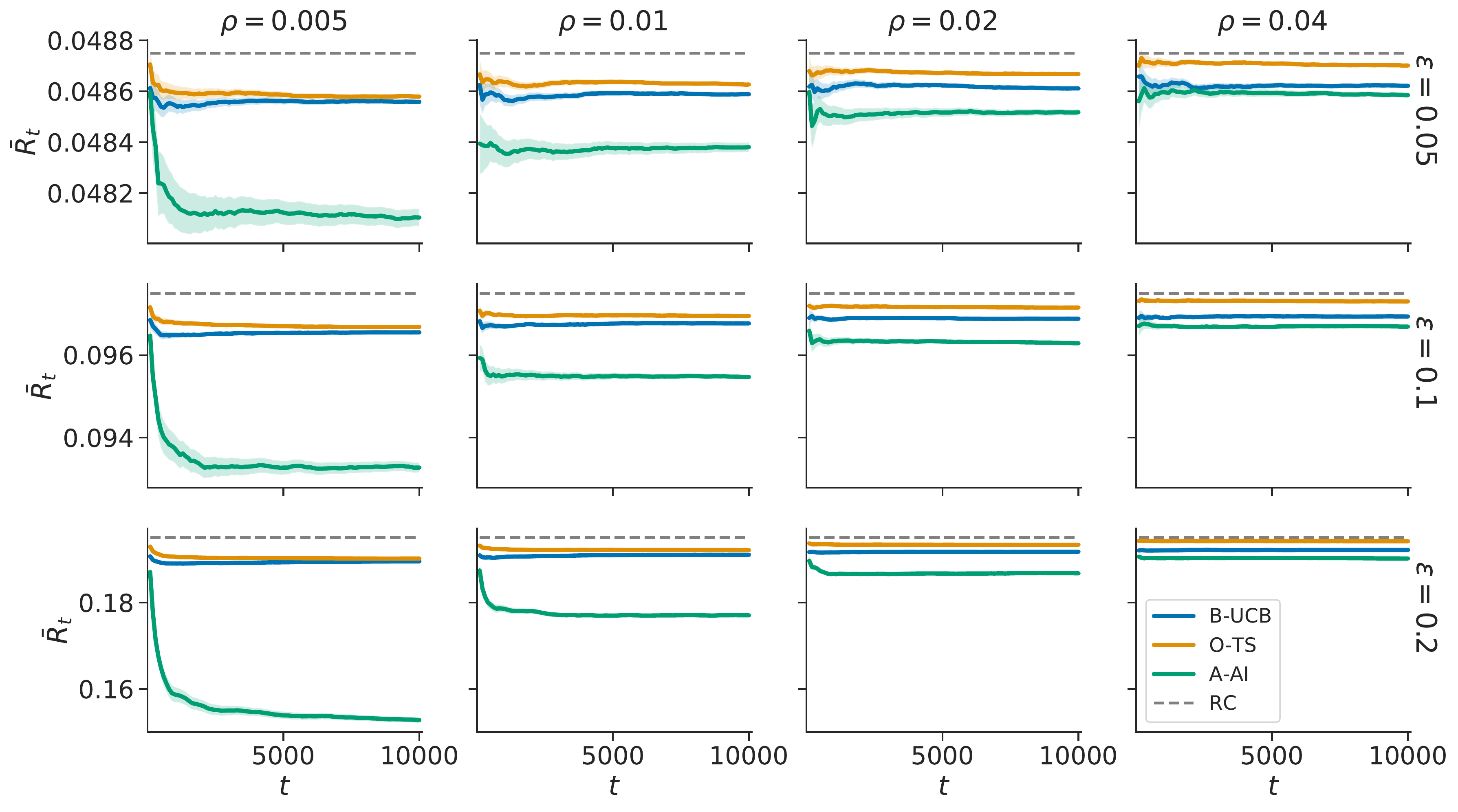}
\caption{{\bf Between-agent comparison in the switching Bernoulli bandit with a fixed number of arms ($K=40$)}. Comparison of the regret rate of approximate active inference (A-AI), optimistic Thompson sampling (O-TS), and Bayesian upper-confidence-bound (B-UCB) based agents in the switching bandit (see \nameref{sec:switching} subsection). Each column and row of the plot corresponds to different task difficulties, characterised by the change probability $\rho$, and the mean outcome difference $\epsilon$. For the A-AI agents the prior precision over outcome preferences is fixed to  $\lambda=0.5$. As in the previous figure, all the values are estimated as ensemble averages over $N=10^3$ simulations, with instance specific switching schedule within the ensemble. The 95\% confidence intervals, although plotted, are in most cases not larger then the line thickness.}
\label{fig:dyn_comp_diff}
\end{figure}

In the case of our switching bandit problem, the change probability $\rho$ acts as an additional difficulty parameter, besides the number of arms $K$ and the mean outcome difference $\epsilon$. Therefore, for the between-algorithm comparison we will first keep $\epsilon$ fixed at its medial value, $\epsilon=0.1$ and vary number of arms in \cref{fig:dyn_comp_arms}, and then keep the number of arms fixed at $K=40$ and vary the expected outcome difference in \cref{fig:dyn_comp_diff}. \chg{lambda2}{For the algorithm comparison in switching bandits we fix the precision parameter $\lambda$, to $\lambda=0.5$, based on a similar visual inspection of regret rate dependence on $\lambda$ (see \cref{fig:sup_comp1}). Interestingly, in the case of switching bandits the minimum of the regret rate stabilises after certain trial number and is not dependent on $T$, like in stationary case}. Note that in stationary environments small values of $\lambda$ are desirable to achieve low cumulative regret for large $T$, in switching environments larger values of $\lambda$ are preferable. \chg{q}{Furthermore, we find that the larger the arm number $K$ is, the larger would be the preferable $\lambda$ value. However, here we will not optimise $\lambda$ for different difficulty settings but use the same value in all examples}. For between-algorithm comparison in switching bandits we will use regret rate, instead of cumulative regret, as a reference performance measure. The reason for this is that in dynamic environments cumulative regret increases linearly with trial number $t$, and regret rate provides visually more accessible gauge of performance differences \cite{raj2017taming}.

In \cref{fig:dyn_comp_arms} we illustrate the regret rate for each agent type over the course of the experiment for a range of different values of change probability $\rho$ and number of arms $K$, and a fixed mean outcome difference $\epsilon=0.1$. Importantly, when estimating the mean regret rate over an ensemble of $N=10^{3}$ agents, for each agent $n \in \cbr{1, \ldots, N}$ we simulate a distinct switching schedule with the same change probability $\rho$. Hence, the average is performed not only over different choice outcome trajectories but also over different hidden trajectories of changes. This ensures that comparison is based on environmental properties, and not on specific realisation of the environment. We find better performance for the active inference agents compared to other bandit algorithms in all conditions. However, we observe that the more difficult the task is (in terms of higher change probability $\rho$ and larger number of arms $K$) the less pronounced is the performance advantage of the active inference based agents.

\begin{figure}[ht!]
\centering
\includegraphics[width=\textwidth]{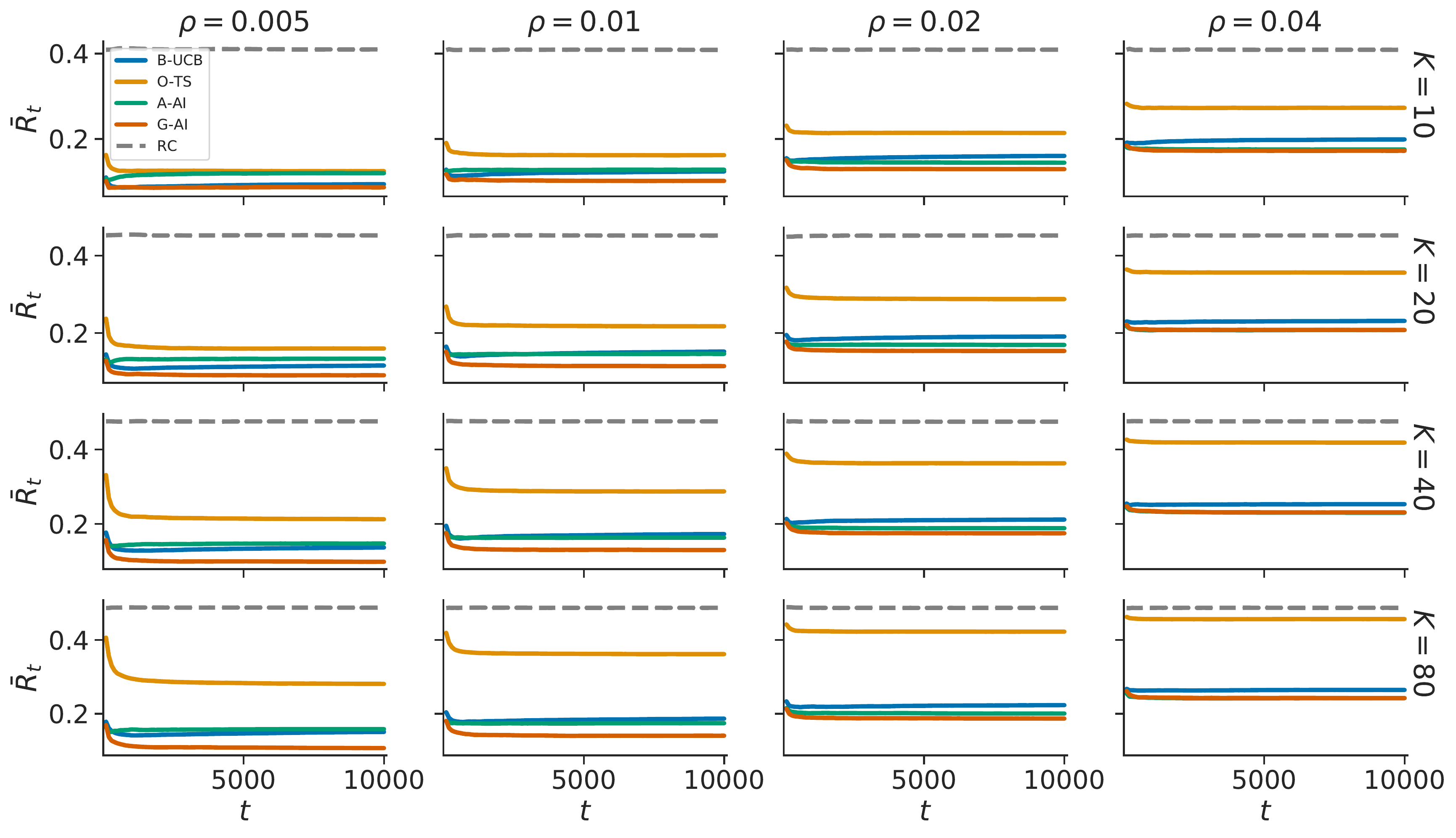}
\caption{{\bf Between-agent comparison in the switching bandit with non-stationary difficulty}. Comparison of the regret rate of exact (G-AI) and approximate active inference (A-AI), optimistic Thompson sampling (O-TS), and Bayesian upper-confidence-bound (B-UCB) agents in the switching bandit (see \nameref{sec:switching} subsection) when the reward probabilities are sampled from uniform distribution after every switch. The re-sampling of latent reward probabilities makes the difficulty of the problem non-stationary, as the advantage of the best arm over the second best arm changes with time. For the A-AI agent we fixed the prior precision over outcome preferences to $\lambda=0.5$, as in the previous examples. However, for the G-AI agent we fixed the lambda to $\lambda=0.25$ based on a visual inspection of dependency of regret rate on $\lambda$ shown in \cref{fig:sup_comp2}. All the values are estimated as ensemble averages over $N=10^3$ simulations, and result in tight confidence intervals.}
\label{fig:dyn_comp_nonstat}
\end{figure}

In \cref{fig:dyn_comp_diff} we show the regret rate for each agent type, however with a fixed number of arms, $K=40$, but varying mean outcome difference $\epsilon$. Here, the picture is very similar, where for increasing task difficulty the A-AI agent type exhibits a diminishing performance advantage relative to the bandit algorithms. Importantly, although we present here the regret analysis only up to $T=5 \cdot 10^3$, unlike in the stationary bandit problem, the results do not change after a further increase in the number of trials. When we simulate longer experiments we find a convergent performance for all algorithms towards a non-zero regret rate; implying a linear increase in cumulative regret with trial number $t$.
Finally, we further illustrate the dependence of performance on mean outcome preference, using the switching bandit with non-stationary task difficulty, where $\epsilon$ is not fixed but changes stochastically over the course of experiment (see \nameref{sec:switching} for more details).
\chg{r}{Importantly, in the case of non-stationary difficulty we will include the G-AI algorithm into comparison. The regret rate based comparison between A-AI and G-AI algorithms (shown in \cref{fig:sup_comp2}) reveals, for the first time, a noticeable difference between the two algorithms. Furthermore, the G-AI algorithm shows a more stable minimum of the regret rate as a function of $\lambda$ in a range of conditions, corresponding to 
$\lambda=0.25$, suggesting potential benefits of exact form of active inference over the approximate one.} As shown in \cref{fig:dyn_comp_nonstat}, we find an increasing advantage of G-AI (and similarly A-AI algorithm) over B-UCB and O-TS algorithms (\cref{fig:dyn_comp_arms}) in more difficult problems -- with either larger number of arms $K$ or larger change probability $\rho$. However, the opposite is the case for lowered task difficulty; {\it e.g.}  for $\rho=0.005$ and $K=10$, where B-UCB achieves higher performance then A-AI algorithm, but is matched with the G-AI algorithm. Notably, we would expect that for small number of arm ($K<10$) and slower changing environments ($\rho<0.001$) the drop in performance of the AI agents becomes even more pronounced, as we are approaching the stationary limit. 

As a final remark, we find it interesting that the B-UCB algorithm consistently outperforms the O-TS algorithm, in almost all non-stationary problems we examined. This is in contrast to the previous asymptotic analysis in the stationary bandit problem, which concluded that Thompson sampling exhibits better asymptotic scaling than B-UCB \cite{kaufmann2012thompson,kaufmann2018bayesian}. We are not aware of previous works comparing these two algorithms in the context of the switching bandit problem. \chg{bucb-guess}{However, the two papers which we found to compare B-UCB and TS in stationary bandits \cite{kaufmann2018bayesian,kaufmann2012thompson} show similar patterns in cumulative regret to what we found. For the initial $T=1,000$ trials B-UCB achives lower regret, and TS outperforms B-UCB only at later stages. Hence, we would infer from these findings that in the switching case B-UCB achieves a lower regret rate as changes occur on a shorter time scale, similar to the advantage we find for the A-AI and G-AI algorithms.}

% --------------------------------------------------------------------------------
% Discussion
% --------------------------------------------------------------------------------

\section{Discussion}
\label{sec:discussion}

In this paper we provide an empirical comparison between active inference, a Bayesian information-theoretic framework \cite{friston2017process}, and two state-of-the-art machine learning algorithms -- Bayesian UCB and optimistic Thompson sampling -- in stationary and non-stationary stochastic multi-armed bandits. We introduced an approximate active inference algorithm, for which our checks on the stationary bandit problem showed that its performance closely follows that of the exact version. Hence, we derived an active inference algorithm that is efficient and easily scalable to high-dimensional problems. 

To our surprise, the empirical algorithm comparison in the stationary bandit problem showed that the active inference algorithm is not asymptotically efficient -- the cumulative regret increased faster than logarithmic in the limit of large number of trials. The cause for this behaviour seems to be the fixed prior precision over preferences $\lambda$, which acts as a balancing parameter between exploration and exploitation. An analysis of how the performance depends on this parameter showed that parameter values that give the best performance decrease over time, suggesting that this parameter should be adaptive and decay over time as the need for exploration decreases. Attempts to remedy the situation with a simple and widely used decay scheme were not successful (for example logarithm of time, not reported here). \chg{adaptive-lambda}{Similarly, introducing a hyper-prior over $\lambda$ and deriving learning rules for the precision parameter \cite{da2020active,sajid2019active} did not result in the desired asymptotic behaviour.} This indicates it is not a simple relationship and a proper theoretical analysis will be needed to identify whether such a scheme exists.

In the non-stationary switching bandit problem the active inference algorithm generally outperformed Bayesian UCB and optimistic Thompson sampling. This provides evidence that the active inference framework may provide a good solution for optimisation problems that require continuous adaptation. Active inference provides the most efficient way of gaining information and this property of the algorithm pays off in the non-stationary setting. Such dynamic settings are also relevant in neuroscience, as relevant changes in choice-reward contingencies are typically hidden and stochastic in everyday environments of humans and other animals \cite{schulz2019algorithmic, gottlieb2013information, wilson2012inferring, cohen2007should, behrens2007learning}. In contrast to previous neuroscience research that showed that active inference is a good description of human learning and decision making \cite{limanowski2020active,adams2020variability,smith2020imprecise,cullen2018active,schwartenbeck2015evidence,schwartenbeck2015dopaminergic}, our results on the dynamic switching bandit show that active inference also performs well in objective sense. Such explanations of cognitive mechanisms that are grounded in optimal solutions are arguably more plausible \cite{chater1999ten}. Hence, this result lends additional credibility to active inference as a generalised framework for understanding human behaviour, not only in the behavioural experiments inspired by multi-armed bandits \cite{izquierdo2017neural,markovic2019predicting,iglesias2013hierarchical,wilson2012inferring,racey2011pigeon,behrens2007learning}, but in a range of related investigations of human and animal decision making in complex dynamic environments under uncertainty \cite{adams2020variability,markovic2020meta,adams2013predictions,pezzulo2012active}. 

An important next step in examining active inference in the context of multi-armed bandits is to establish theoretical bounds on the cumulative regret for the stationary bandit problem. A key part of these theoretical studies will be to investigate whether it is possible to devise a sound decay scheme for the $\lambda$ parameter (see \pcref{eq:approx_ai}), that provably works for all instances of the canonical stationary bandit. This would lead to the development of new active inference inspired algorithms which can achieve asymptotic efficiency. These theoretical bounds would allow us to more rigorously compare active inference algorithms to the already established bandit algorithms for which regret bounds are known. Moreover, we would potentially be able to generalise beyond the settings we have empirically tested here. Future work may also consider an information-theoretic analysis of active inference, which might be more appropriate than regret analysis \cite{russo2018tutorial}. For example, the Bayesian exploration bonus previously considered in Bayesian reinforcement learning was analysed with respect to sample complexity of identifying a good policy \cite{kolter2009near}. Similarly, in \cite{russo2016information} the authors introduced a new measure of regret weighted by the inverse information gain between actions and outcomes, and provided expected bounds for this measure for several Bayesian algorithms, such as Thompson sampling and Bayesian UCB. \chg{s}{Finally, future work should contrast the active inference framework with alternative approaches that can generate directed exploration \cite{russo2018learning,frazier2008knowledge}.}

As optimal behaviour is always defined with respect to a chosen objective function, a different objective function will lead to different behaviour, and the appropriateness of the objective function for the specific problem determines the performance of the algorithm on a given task. In other words, behaviour is determined not only by the beliefs about the hidden structure about the states of the world but also by the beliefs about useful preferences and objectives one should take into account in that environment. Therefore, although one can consider the sensitivity of the introduced active inference algorithm on the prior precision over preferences $\lambda$ as a limitation of the algorithm in comparison to the other two algorithms, we believe that it is possible to introduce various adaptations to the algorithm to improve asymptotic behaviour. Fore example, one can consider learning rules for prior outcome preferences, as illustrated in \cite{sajid2019active,markovic2020meta}. This would introduce a way to adapt an objective function to different environments achieving high performance in a wide-range of multi-armed bandit problems. Alternatively, instead of basing action selection on the expected free energy, one can define a stochastic counterpart, which is estimated based on samples from the posterior, akin to Thompson sampling. This would enable the algorithm to better leverage directed and random exploration. 

Despite of the poor asymptotic performance in the stationary bandit problem there are some advantages of active inference over classical bandit algorithms, both for artificial intelligence and neuroscience. Unlike the Thompson sampling and UCB algorithms, active inference is easily extendable to more complex settings where actions affect future states and actions available. Such settings are usually formalised as a (partially observable) Markov decision process, which require the combination of adaptive decision making with complex planning mechanism \cite{millidge2020deep,ueltzhoffer2018deep,fountas2020deep}. In these settings learning is non-stationary because changes in policy cause a shift in state value distributions \cite{sutton2018reinforcement}. Given our finding that active inference algorithm has an advantage in non-stationary settings, it seems promising to apply the framework to Markov decision processes. Reinforcement learning algorithms is a popular choice for tackling Markov decision processes, in particular it would be interesting to compare active inference to Bayesian reinforcement learning approaches \cite{ghavamzadeh2015bayesian,guez2013scalable,guez2014bayes}.

The generative modelling approach integral to active inference allows several improvements to the presented algorithm, which also holds for related Bayesian approaches. For example, we have considered here only one learning algorithm, variational SMiLE \cite{liakoni2021learning}, which we have chosen based on its simplicity and efficiency. A potential drawback of variational SMiLE is that it might not be optimal (in terms of inference) for \chg{t}{switching bandits or} a generic problem of dynamic bandits (e.g. different mechanisms for generating changes and different reward distribution). For example, \chg{alternative-switching-learning}{for switching bandits several candidates come closer to the exact inference and would likely improve performance \cite{adams2007bayesian,mellor2013thompson,lu2019adaptive,alami2020restarted}.} For restless bandits, which follow a random walk process, recently published alternative efficient learning algorithms derived from different generative models are likely to provide a better performance \cite{piray2020simple,moens2019learning}. Employing a good learning algorithm is especially important in dynamic settings, where exact inference is not tractable, and the performance of learning rules is tightly coupled to the overall performance of the algorithm. In practice, one would expect that the better the generative model and the corresponding approximate inference algorithm, the better the performance will be on a given multi-armed bandit problem. Furthermore, one can easily extend the learning algorithms with deep hierarchical variants, which can infer a wide range of unknown dynamical properties of the environment \cite{piray2020simple} and learn higher order temporal statistics \cite{markovic2019predicting, markovic2020meta}.

% --------------------------------------------------------------------------------
% Conclusion
% --------------------------------------------------------------------------------

\section{Conclusion}

We have derived an approximate active inference algorithm, based on a Bayesian information-theoretic framework recently developed in neuroscience, proposing it as a novel machine learning algorithm for bandit problems that can compete with state-of-the-art bandit algorithms. Our empirical evaluation has shown that the active inference framework can indeed be used to derive a promising bandit algorithm. We consider the present work as a first step, where two important next steps are the development of a decay schedule for the outcome preference precision parameter $\lambda$ and a theoretical regret analysis for the stationary bandit. The fact that the active inference algorithm achieves excellent performance in switching bandit problems, commonly used in cognitive neuroscience, provides rational grounds for using active inference as a generalised framework for understanding human and animal learning and decision making. 

\section{Acknowledgements}

We thank Karl Friston and Gergely Neu for valuable feedback and constructive discussions. DM and SS were funded by the German Research Foundation (DFG, Deutsche Forschungsgemeinschaft), SFB 940/3 - Project ID 178833530, A09,TRR 265/1 - Project ID 402170461, B09 and partially supported by Germany's Excellence Strategy – EXC 2050/1 – Project ID 390696704 – Cluster of Excellence ``Centre for Tactile Internet with Human-in-the-Loop'' (CeTI) of Technische Universität Dresden. The Max Planck UCL Centre for Computational Psychiatry and Ageing Research is funded by the Max Planck Society, Munich, Germany, URL: \url{https://www.mpg.de/en}, grant number: 647070403019.

% --------------------------------------------------------------------------------
% Bibliography
% --------------------------------------------------------------------------------

\bibliographystyle{elsarticle-num}
\bibliography{references}

% --------------------------------------------------------------------------------
% Appendix
% --------------------------------------------------------------------------------

\appendix
\section*{Appendix}
\label{sec:appendix}

\addcontentsline{toc}{section}{Appendices}
\renewcommand{\thesubsection}{\Alph{subsection}}
\renewcommand{\theequation}{\thesubsection.\arabic{equation}}

\subsection{Deriving the expression for the expected free energy}
\label{sec:derive_efe}

We write the approximate (exact in the case of stationary Bernoulli bandits) posterior beliefs about reward probabilities $\vec{\theta}_{t-1}$ at trial $t-1$ as
\begin{equation}
    Q\del{\vec{\theta}_{t-1}|\vec{\eta}_t} = \prod_{k=1}^K \mathcal{B}e\del{\alpha_{t-1, k}, \beta_{t-1, k}},
\end{equation}
where $\vec{\eta}_{t-1} = (\alpha_{t-1, 1}, \beta_{t-1, 1}, \ldots, \alpha_{t-1, K}, \beta_{t-1, K})$, contains the information about the history of choices $a_{1:t-1}$ and 
outcomes $o_{1:t-1}$ up to trial $t-1$. Next, we obtain the predictive prior distribution at trial $t$ as 
\begin{equation}
\begin{split}
    p\del{\vec{\theta}_{t}|j_t, \vec{\eta}_{t-1}} &= \int \dif \vec{\theta}_{t-1} p\del{\vec{\theta}_{t}| \vec{\theta}_{t-1}, j_t} Q\del{\vec{\theta}_{t-1}|\vec{\eta}_t} \\ &= \left\{ \begin{array}{ll} \prod_{k=1}^K \mathcal{B}e\del{\alpha_{t-1, k}, \beta_{t-1, k}} & \text{for } j_t = 0 \\ \prod_{k=1}^K \mathcal{B}e\del{\alpha_{0,k}, \beta_{0,k}} & \text{for } j_t = 1 \end{array}\right.
\end{split}
\end{equation}
where $\alpha_{0,k} = \beta_{0,k} = 1$. Marginalising out $j_t$ from the joint predictive distribution $p\del{\vec{\theta}_{t}|j_t, \vec{\eta}_{t-1}}$ leads to the following marginal predictive probability 
\begin{equation}
    p\del{\vec{\theta}_{t}|\vec{\eta}_{t-1}} = (1 - \rho) \prod_{k=1}^K \mathcal{B}e\del{\alpha_{t-1,k}, \beta_{t-1,k}} + \rho \prod_{k=1}^K \mathcal{B}e\del{\alpha_{0,k}, \beta_{0,k}}.
\end{equation}
Finally we compute the probability of observing outcome $o_t$ given action $a_t$ on current trial $t$ by marginalising the joint distribution $p\del{o_t, \vec{\theta}_{t}|a_t, \vec{\eta}_{t-1}}$ over latent states $\vec{\theta}_t$. Hence,
\begin{equation}
\begin{split}
    Q\del{o_t|a_t, \vec{\eta}_{t-1}} &= \int \dif \vec{\theta}_t p\del{o_t|a_t, \vec{\theta}_t} p\del{\vec{\theta}_{t}|\vec{\eta}_{t-1}} \\ 
    &= \prod_{k=1}^K \sbr{\tilde{\mu}_{t,k}^{o_t}\del{1-\tilde{\mu}_{t,k}}^{1-o_t}}^{\delta_{a_t,k}},
\end{split}
\end{equation}
where $\tilde{\mu}_{t, k}$ corresponds to the expression used in \cref{eq:marginal}. To compute the expected free energy $G(a_t)$ we will split the full expression on two terms
the risk, denoted with $G_R(a_t)$, and the ambiguity, denoted with $G_A(a_t)$. Hence, 
\begin{equation}
    G(a_t) = G_R(a_t) + G_A(a_t).
\end{equation}
Combining \cref{eq:preference} and \cref{eq:marginal} we compute the risk using the following relation

\begin{equation}
\begin{split}
        G_R(a) &= D_{KL}\del{Q(o_t |a_t = a)||P(o_t)} \\
        &= \sum_{o_t} Q(o_t|a) \ln \frac{Q(o_t|a)}{P(o_t)} \\
        &= const. + \sum_{o_t} Q(o_t|a) \sbr{ o_t \del{ \ln \tilde{\mu}_{t, a} - \lambda} + (1 - o_t) \del{ \ln \del{1- \tilde{\mu}_{t, a}} + \lambda} } \\
        & = const. + \sbr{ \tilde{\mu}_{t, a} \del{ \ln \tilde{\mu}_{t, a} - \lambda} + (1 - \tilde{\mu}_{t, a}) \del{ \ln \del{1- \tilde{\mu}_{t, a}} + \lambda} } \\
        & = - \lambda \del{2 \tilde{\mu}_{t, a} - 1} + \tilde{\mu}_{t, a} \ln \tilde{\mu}_{t, a} + (1 - \tilde{\mu}_{t, a}) \ln \del{1- \tilde{\mu}_{t, a}} + const.
\end{split}
\label{eq:risk}
\end{equation}

To derive the expression for the ambiguity part $G_A(a)$ of the expected free energy we will start by computing conditional entropy of outcomes $o_t$, defined as 
\begin{equation}
\begin{split}
    H\sbr{o_t|\vec{\theta}_t, a_t} &= - \sum_{o_t} p(o_t|\vec{\theta}_t, a_t) \ln p(o_t|\vec{\theta}_t, a_t) \\
    & = - \sum_{o_t} p(o_t|\vec{\theta}_t, a_t) \sum_{k=1}^K \delta_{a_t, k} \sbr{o_t \ln \theta_{t,k} + (1-o_t) \ln \del{1-\theta_{t,k}}} \\
    & = - \theta_{t, a_t} \ln \theta_{t, a_t} - \del{1 - \theta_{t, a_t}}\ln \del{1-\theta_{t,k}}.
\end{split}
\end{equation}

As the ambiguity term corresponds to expectation over latent states of the conditional entropy we obtain the following expression
\begin{multline}
    G_A(a) = E_{p(\vec{\theta}_t|\vec{\eta}_{t-1})} \sbr{H\sbr{o_t|\vec{\theta}_t, a_t=a}} \\
    =  - (1-\rho)\sbr{\mu_{t-1, a} \psi(\alpha_{t-1,a}) + (1-\mu_{t-1, a}) \psi(\beta_{t-1, a}) - \psi(\nu_{t-1, a}) + \frac{1}{\nu}}  + const.
    \label{eq:ambi}
\end{multline}

Where we used the following relations 
\begin{equation}
\begin{split}
    \int d x \mathcal{B}e\del{x; \alpha, \beta} x \ln x \\
    &= \frac{B(\alpha+1, \beta)}{B(\alpha, \beta)} \int d x B\del{x; \alpha+1, \beta} \ln x \\
    & = \frac{B(\alpha+1, \beta)}{B(\alpha, \beta)} \sbr{\psi(\alpha+1) - \psi\del{\nu+1}} \\
    &= \mu \sbr{\psi(\alpha+1) - \psi\del{\nu+1}} \\
    & =  \mu \sbr{\psi(\alpha) + \frac{1}{\alpha} - \psi\del{\nu} - \frac{1}{\nu}} \\
    & = \mu \psi(\alpha) + \frac{1}{\nu} - \mu \sbr{ \psi\del{\nu} + \frac{1}{\nu}}
\end{split}
\end{equation}
and
\begin{equation}
\begin{split}
    \int d x \mathcal{B}e\del{x; \alpha, \beta} (1-x) \ln (1-x) \\
    &= \frac{B(\alpha, \beta+1)}{B(\alpha, \beta)} \int d x B\del{x; \alpha, \beta+1} \ln (1-x) \\
    & = \frac{B(\alpha, \beta+1)}{B(\alpha, \beta)} \sbr{\psi(\beta+1) - \psi\del{\nu+1}}\\    
    &= (1-\mu) \sbr{\psi(\beta+1) - \psi\del{\nu+1}} \\
    & =  (1-\mu) \sbr{\psi(\beta) + \frac{1}{\beta} - \psi\del{\nu} - \frac{1}{\nu}} \\
    & = (1-\mu) \psi(\beta) + \frac{1}{\nu} - (1-\mu) \sbr{ \psi\del{\nu} + \frac{1}{\nu}}
\end{split}
\end{equation}
to obtain the expectations of conditional entropy. Combining \cref{eq:risk,eq:ambi} we get the expression for the expected free energy of action $a$ on trial $t$ shown in \cref{eq:efe_est}.

\subsection{Deriving the approximate expression of the expected free energy}
\label{sec:derive_app}

To derive the approximate expression for the expected free energy shown in \cref{eq:approx_ai} we note that for a sufficiently large $x$ the following relation holds \cite{bernardo1976algorithm} 
\begin{equation}
    \psi(x) = \ln x - \frac{1}{2x}.
\end{equation}
As parameters of the Beta distribution are monotonically increasing with each update, we can assume that they reach large enough value after certain number of trials. Hence, instead of using the exact expression for ambiguity derived in \cref{eq:ambi}, we can used a simplified expression based on the mentioned approximation of the digamma function. 
Therefore, the approximate ambiguity term becomes
\begin{equation}
\begin{split}
    G_A(a) &\approx  - (1-\rho) \mu_{t-1, a} \del{ \ln\alpha_{t-1,a} - \frac{1}{2\alpha_{t-1,a}} } \\ 
    & \quad - (1-\rho) (1-\mu_{t-1, a}) \del{ \ln\beta_{t-1, a} - \frac{1}{2\beta{t-1, a}}} \\
    & \quad + (1-\rho) \del{ \ln\nu_{t-1, a} - \frac{3}{2\nu}}  + const. \\
    & = - (1-\rho) \sbr{\mu_{t-1, a} \ln \mu_{t-1, a} + \del{1 - \mu_{t-1, a}} \ln \del{1 - \mu_{t-1, a}} + \frac{1}{2 \nu_{t-1, a}} }
\end{split}
\end{equation}

As we are interested only in cases for which change probability is small, namely $\rho \leq 0.1$ we can further approximate the risk term of the expected free energy as
\begin{equation}
\begin{split}
    G_R(a) &\approx - (1 - \rho) \sbr{ 2 \lambda \mu_{t-1, a} -  \mu_{t-1, a} \ln \tilde{\mu}_{t, a} - \del{1 -  \mu_{t-1, a}} \ln \del{1 - \tilde{\mu}_{t,a}} } \\
    &\approx - (1 - \rho) \sbr{ 2 \lambda \mu_{t-1, a} -  \mu_{t-1, a} \ln \mu_{t-1, a} - \del{1 -  \mu_{t-1, a}} \ln \del{1 - \mu_{t-1,a}} }
\end{split}
\end{equation}
where we set $\ln \tilde{\mu}_{t, a} \approx \ln \mu_{t-1, a}$ and $\ln \del{1 - \tilde{\mu}_{t, a}} \approx \ln \del{ 1- \mu_{t-1, a}}$. Adding together the two approximate terms leads to the following expression for the approximate free energy
\begin{equation}
    G_t(a)\approx - (1 - \rho) \sbr{ 2 \lambda \mu_{t-1, a} + \frac{1}{2 \nu_{t-1, a}} }.
\end{equation}
Thus minimising approximate expected free energy corresponds the expression shown in \cref{eq:approx_ai}, where cancelling the negative sign turns minimisation into maximisation process.

\subsection{Benchmark of action selection algorithms}
\label{sec:benchmark}

All multi-armed bandits algorithms used in this paper have been implemented using JAX (Autorgrad and XLA) \cite{jax2018github} and executed in Python 3.9.4 environment. JAX uses XLA (Accelerated Linear algebra) to just-in-time compile Python functions into XLA-optimized kernels and run them on GPUs and TPUs. For the benchmarks presented in \cref{tab:benchmark} we have used a Lenovo Workstation with AMD Threadripper 3955WX CPU, NVIDIA Quadro RTX 4000 GPU, and 64 GB RAM. 

\begin{table}
\centering
\begin{tabular}{|c|c|c|c|c|}
\hline 
Algorithm & $K=10$ & $K=20$ & $K=40$ & $K=80$ \\ 
\hline 
UCB & 0.038 & 0.039 & 0.042 & 0.042 \\ 
\hline 
B-UCB & 0.963 & 0.980 & 1.052 & 1.185 \\ 
\hline 
TS & 0.992 & 0.967 & 1.207 & 1.864 \\ 
\hline 
O-TS & 0.968 & 0.939 & 1.034 & 1.609 \\ 
\hline 
G-AI & 0.041 & 0.041 & 0.054 & 0.072 \\ 
\hline
A-AI & 0.034 & 0.036 & 0.039 & 0.040 \\ 
\hline
\end{tabular}
\caption{Compute times per decision presented in milliseconds. The compute times were estimated as an average over ten repetitions of a $T=10000$ long loop, consisting of only action selection and learning algorithm. Outcomes were kept fixed on all trials. Each algorithm was executed with $N=1000$ parallel simulations. For comparison, we show run times of classical variants of UCB and TS \cite{chapelle2011empirical}.}
\label{tab:benchmark}
\end{table}

Note that the presented compute times can hardly be generalised to other multi-armed bandit problems, and are highly dependent on the efficiency of JAX framework to optimise various functions, and sampling algorithms. 

\subsection{Supplementary Figures}

\renewcommand{\thefigure}{S\arabic{figure} }
\setcounter{figure}{0}

\begin{figure}[ht!]
\centering
\includegraphics[width=\textwidth]{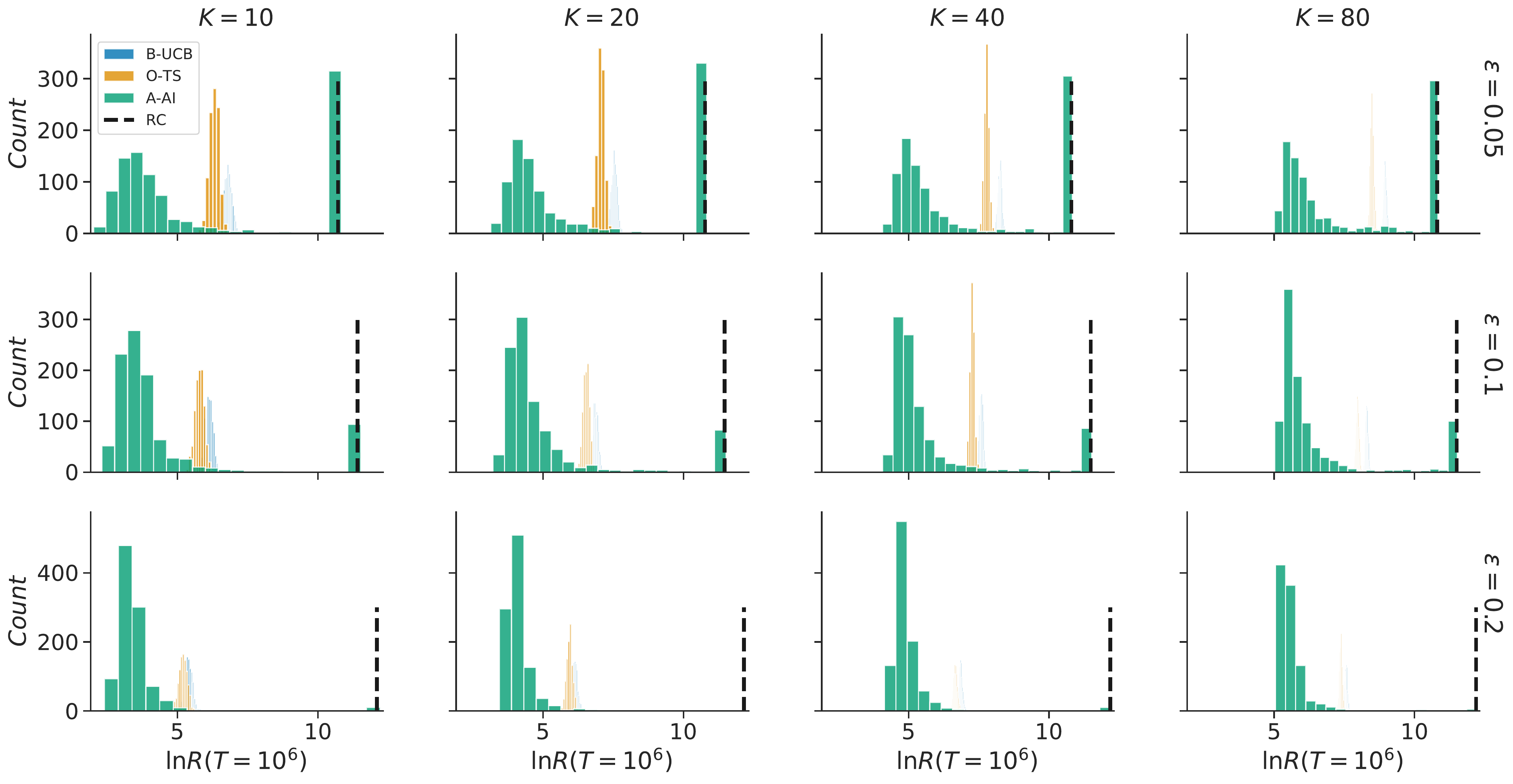}
\caption{{\bf Histogram of the logarithm of cumulative regret}. The ensemble based distribution the cumulative regret at $T=10^6$ for different algorithms estimated from $N=10^3$ simulations. Note that the peak in the tail of the distribution for A-AI algorithm, proportional to random choices, implies that percentage of agents in the ensemble never found a correct solution. Hence, as number of trials increases the average cumulative regret over the ensemble is pulled towards values which grow linearly with the trial number $t$.}
\label{fig:sup_hist}
\end{figure}

\begin{figure}[ht!]
\centering
\includegraphics[width=\textwidth]{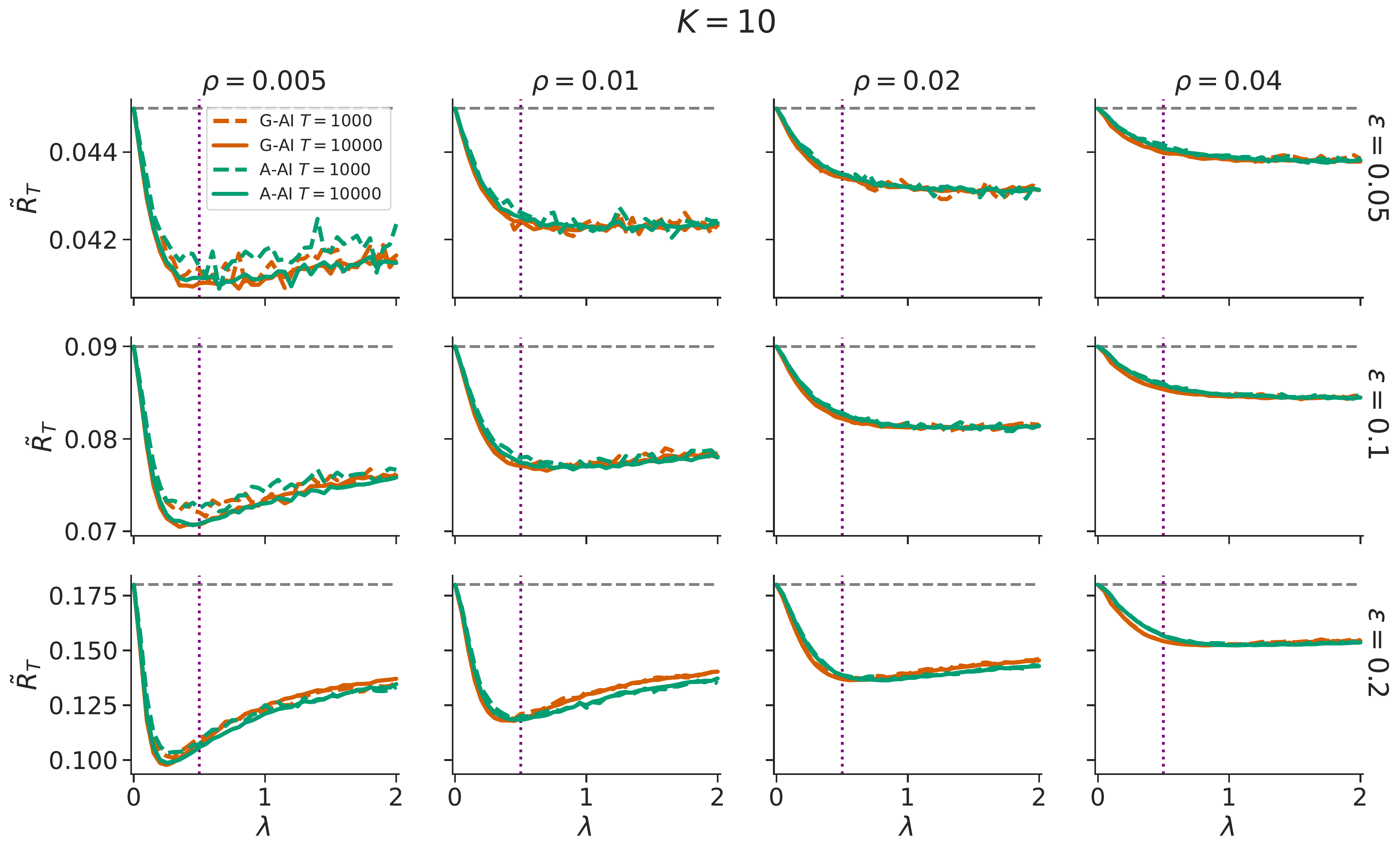}
\caption{{\bf Regret rate analysis for active inference based agents in the switching Bernoulli bandits with fixed difficulty}. The regret rate $\tilde{R}_T$, \cref{eq:reg_rate}, for the approximate (A-AI) and the exact (G-AI) variants of active inference as a function of the precision over prior preferences $\lambda$. The coloured lines show numeric estimates obtained as an average over $N=10^3$ runs. Different line styles denote $\tilde{R}_T$ values estimated after different numbers of trials $T$. The dashed black line denotes the upper bound on the regret rate corresponding to the random (RC) agent which gains no information from the choice outcomes. The vertical dotted line (purple) corresponds to $\lambda=0.5$, which we find to be sufficiently close to the minimum or regret rate in a range of conditions. Each column and row of the plot corresponds to different task difficulties, characterised by the change probability $\rho$, and the mean outcome difference $\epsilon$, respectively. We have fixed the arm number to $K=10$.}
\label{fig:sup_comp1}
\end{figure}

\begin{figure}[ht!]
\centering
\includegraphics[width=\textwidth]{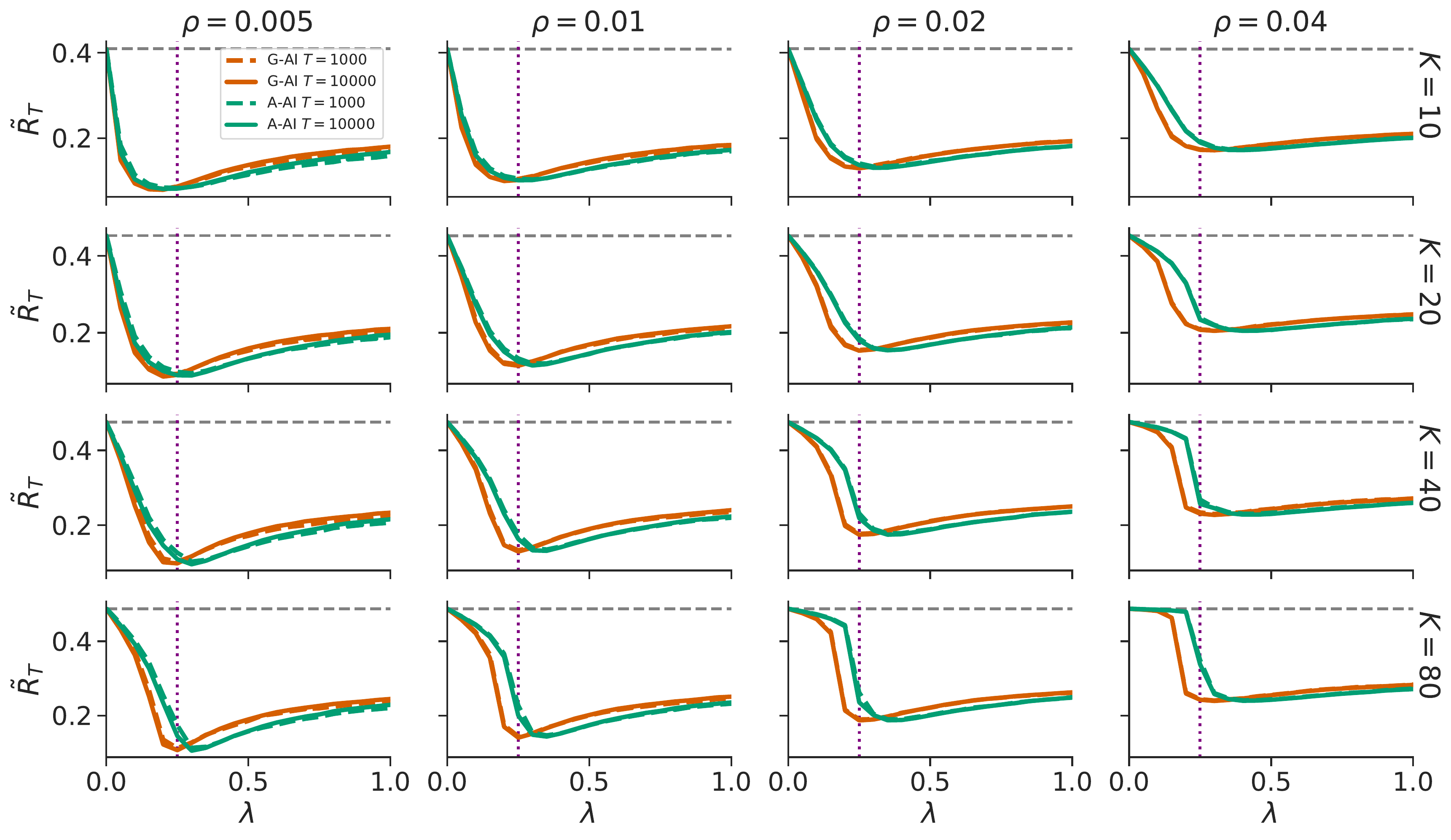}
\caption{{\bf Regret rate analysis for active inference based agents in the switching Bernoulli bandits with varying difficulty}. The regret rate $\tilde{R}_T$, \cref{eq:reg_rate}, for the approximate (A-AI) and the exact (G-AI) variants of active inference as a function of the precision over prior preferences $\lambda$. The coloured lines show numeric estimates obtained as an average over $N=10^3$ runs. Different line styles denote $\tilde{R}_T$ values estimated after different numbers of trials $T$. The dashed black line denotes the upper bound on the regret rate corresponding to the random (RC) agent which gains no information from the choice outcomes. The vertical dotted line (purple) corresponds to $\lambda=0.25$, which we find to be sufficiently close to the minimum or regret rate of the G-AI algorithm in a range of conditions. Each column and row of the plot corresponds to different task difficulties, characterised by the change probability $\rho$, and the arm number $K$, respectively.}
\label{fig:sup_comp2}
\end{figure}

\end{document}